\pdfoutput=1

\documentclass[11pt]{article}

\usepackage[final]{acl}

\usepackage{times}
\usepackage{latexsym}

\usepackage[T1]{fontenc}

\usepackage[utf8]{inputenc}

\usepackage{microtype}

\usepackage{inconsolata}

\usepackage{graphicx}

\usepackage{booktabs}
\usepackage{xcolor,colortbl}
\usepackage{fontawesome5}
\usepackage{pgfplots}
\pgfplotsset{width=10cm,compat=1.9}
\usepackage{tikz}
\usepackage{tcolorbox}
\usepackage{paracol}
\usepackage{makecell}
\usepackage{pifont}
\usepackage{tcolorbox}
\usepackage{supertabular}
\usepackage{subcaption}
\usepackage{import}
\usepackage{multirow}
\usepackage{subcaption}
\usetikzlibrary{patterns}

\newcommand{\name}{\textsc{Faina}}

\definecolor{Gray}{gray}{0.95}
\newcolumntype{a}{>{\columncolor{Gray}}r}
\definecolor{forestgreen}{RGB}{34,139,34}
\definecolor{indianred}{RGB}{205,92,92}
\newcommand{\cmark}{\textcolor{forestgreen}{\ding{51}}}%
\newcommand{\xmark}{\textcolor{indianred}{\ding{55}}}%

\newcommand{\coldface}{\raisebox{-2pt}{\includegraphics[width=1.2em]{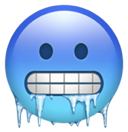}}}
\newcommand{\clappinghands}{\raisebox{-2pt}{\includegraphics[width=1.2em]{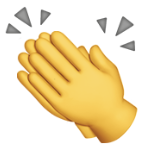}}}
\newcommand{\facewithsteam}{\raisebox{-2pt}{\includegraphics[width=1.2em]{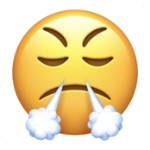}}}
\newcommand{\facewithtears}{\raisebox{-2pt}{\includegraphics[width=1.2em]{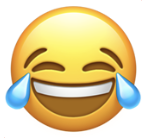}}}
\newcommand{\flagitaly}{\raisebox{-2pt}{\includegraphics[width=1.2em]{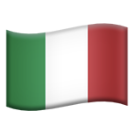}}}
\newcommand{\flexedbiceps}{\raisebox{-2pt}{\includegraphics[width=1.2em]{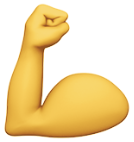}}}
\newcommand{\foldedhands}{\raisebox{-2pt}{\includegraphics[width=1.2em]{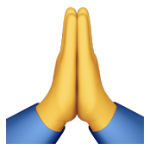}}}
\newcommand{\greenheart}{\raisebox{-2pt}{\includegraphics[width=1.2em]{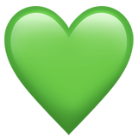}}}

\newcommand{\redheart}{\raisebox{-2pt}{\includegraphics[width=1.2em]{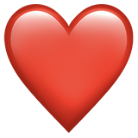}}}
\newcommand{\redsign}{\raisebox{-2pt}{\includegraphics[width=1.2em]{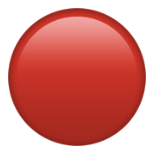}}}
\newcommand{\starr}{\raisebox{-2pt}{\includegraphics[width=1.2em]{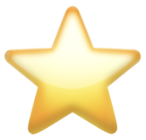}}}
\newcommand{\vaccines}{\raisebox{-3pt}{\includegraphics[width=1.2em]{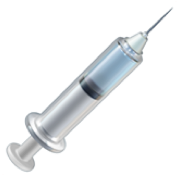}}}
\newcommand{\faina}{\raisebox{-2pt}{\includegraphics[width=1.2em]{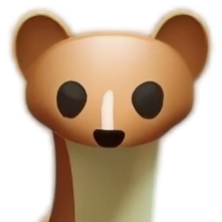}}}

\NewTColorBox{NewBox}{ s O{!htbp} }{%
  floatplacement={#2},
  IfBooleanTF={#1}{float*,width=\textwidth}{float}, colframe=black!50!white,colback=black!2!white,arc=0mm,fonttitle=\bfseries,coltitle=black!2!white,title=\faCog~~Prompt template
}

\title{Fine-grained Fallacy Detection with Human Label Variation}

\author{Alan Ramponi,$^1$ Agnese Daffara,$^{2,3}$ Sara Tonelli$^1$\\ 
\texttt{\{alramponi,satonelli\}@fbk.eu}, \texttt{agnese.daffara@ims.uni-stuttgart.de} \\ \\
\textsuperscript{1} Digital Humanities group, Fondazione Bruno Kessler, Italy \\
\textsuperscript{2} Department of Humanities, University of Pavia, Italy\\
\textsuperscript{3} Institute for Natural Language Processing, University of Stuttgart, Germany}

\begin{document}
\maketitle
\begin{abstract}
We introduce \name, the first dataset for fallacy detection that embraces multiple plausible answers and natural disagreement. \name~includes over 11K span-level annotations with overlaps across 20 fallacy types on social media posts in Italian about migration, climate change, and public health given by two expert annotators. Through an extensive annotation study that allowed discussion over multiple rounds, we minimize annotation errors whilst keeping signals of human label variation. Moreover, we devise a framework that goes beyond ``single ground truth'' evaluation and simultaneously accounts for multiple (equally reliable) test sets and the peculiarities of the task, i.e., partial span matches, overlaps, and the varying severity of labeling errors. Our experiments across four fallacy detection setups show that multi-task and multi-label transformer-based approaches are strong baselines across all settings. We release our data, code, and annotation guidelines to foster research on fallacy detection and human label variation more broadly.\footnote{~\faGithub~Repository:~\url{https://github.com/dhfbk/faina}.}
\end{abstract}


\section{Introduction}

Fallacies are traditionally defined as types of reasoning that seem valid but are not~\citep{hamblin2022fallacies,tindale2007fallacies}. They occur when someone commits an error in the argumentation, either with the purpose of persuading the audience or unintentionally. Disentangling the object of the discussion from the way in which it is expressed is paramount because even true statements can be invalid due to their faulty composition. 
Social media is a perfect ground for studying fallacies, which have moved beyond the Aristotle's realm of two-person debates into the vast domain of the internet. 
Fallacious social media content can mislead a large audience, in some cases leading to the proliferation of misinformation on highly-debated topics~\citep{musi2022developing}. 
Recognizing fallacies in everyday argumentation may not only limit the spread of harmful content but also plays a key role in developing individuals' critical thinking skills, ultimately contributing to mitigate faulty argumentation at its root and promoting democratic debate~\cite{ecker2024misinformation}.

\begin{figure}[t]
  \includegraphics[width=\columnwidth]{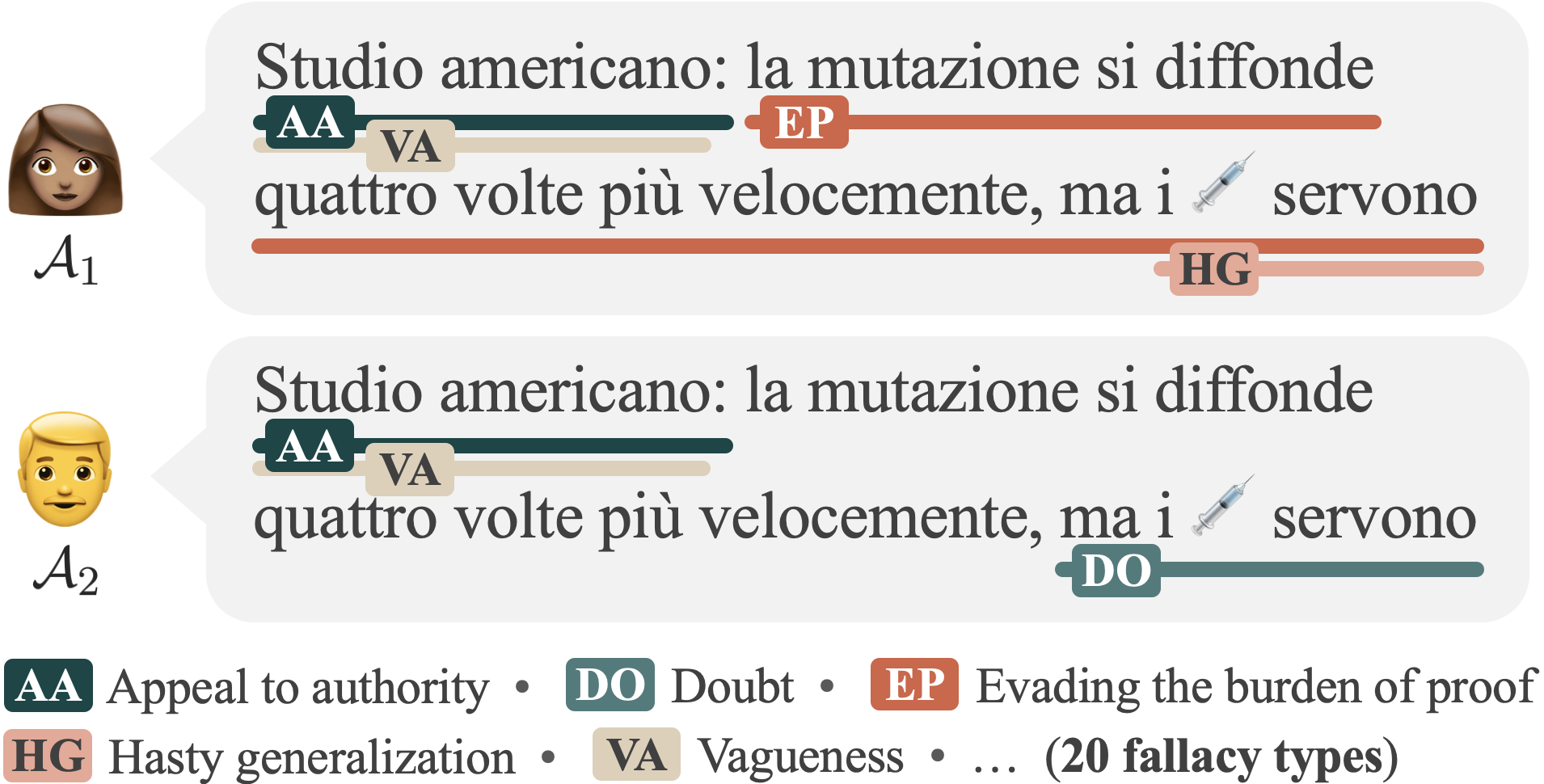}
  \caption{Example showing multiple plausible span annotations provided by annotators $\mathcal{A}_1$ and $\mathcal{A}_2$ due to different interpretations (en: ``\emph{American study: mutation spreads four times faster, but \vaccines~are needed}'').}
  \label{fig:overview}
\end{figure}

\begin{table*}[ht!]
  \centering
  \resizebox{0.99\linewidth}{!}{%
  \begin{tabular}{llllrrcc}
    \toprule
     \textbf{Work} & \textbf{Langs} & \textbf{Genres and domains/topics} & \textbf{Scope} & \textbf{\#C} & \textbf{\#I} & \textbf{Multi} & \textbf{HLV} \\
    \midrule
    Argotario (\texttt{en})~\citep{habernal-etal-2017-argotario} & 
    \texttt{en} & 
    \faBookReader\,~General & 
    pair & 
    5$^\dagger$ &
    909\,\, & 
    \xmark &
    \xmark \\
    Argotario (\texttt{de})~\citep{habernal-etal-2018-adapting} & 
    \texttt{de} & 
    \faBookReader\,~General & 
    pair & 
    5$^\dagger$ &
    335\,\, & 
    \xmark &
    \xmark \\
    ChangeMyView~\citep{habernal-etal-2018-name} & 
    \texttt{en} &
    \faRss\;\;Various topics (discussions) &
    text &
    1$^\dagger$ &
    3,622\,\, & 
    \xmark &
    \xmark \\
    InformalFallacies*~\citep{sahai-etal-2021-breaking} & 
    \texttt{en} &
    \faRss\;\;Politics;\,religion;\,veganism;\,fallacies & 
    span &
    8\,\, &
    1,708\,\, &
    \xmark &
    \xmark \\
    AdHomInTweets~\citep{sheng-etal-2021-nice} & 
    \texttt{en} &
    \faRss\;\;BLM;\,MeToo;\,veganism;\,remote work & 
    pair & 
    6$^\dagger$ &
    2,400$^\ddagger$ & 
    \xmark &
    \xmark \\
    LanguageOfPopulism*~\citep{macagno2022argumentation} & 
    \texttt{en};\,\texttt{it};\,\texttt{pt} &
    \faRss\;\;Political discourse (4 political leaders) & 
    text &
    9$^\dagger$ &
    1,919\,\, &
    \cmark &
    \xmark \\
    Logic~\citep{jin-etal-2022-logical} & 
    \texttt{en} &
    \faBookReader\,~General &
    snippet &
    13\,\, &
    2,449\,\, &
    \xmark &
    \xmark \\
    LogicClimate~\citep{jin-etal-2022-logical} & 
    \texttt{en} &
    \faNewspaper[regular]~Climate change &
    snippet &
    13\,\, &
    1,079\,\, &
    \xmark &
    \xmark \\
    COVID-19~\citep{musi2022developing} & 
    \texttt{en} &
    \faNewspaper[regular]~COVID-19 &
    snippet &
    10\,\, &
    526\,\, &
    \xmark &
    \xmark \\
    Climate~\citep{alhindi-etal-2022-multitask} & 
    \texttt{en} &
    \faNewspaper[regular]~Climate change &
    snippet &
    10\,\, &
    477\,\, &
    \xmark &
    \xmark \\
    ElecDeb60to16~\citep{goffredo2022fallacious} &
    \texttt{en} &
    \faMicrophone\;\:\:Political debates (US pres.~campaigns) &
    snippet & 
    14\,\, &
    1,628\,\, &
    \xmark &
    \xmark \\
    ElecDeb60to20~\citep{goffredo-etal-2023-argument} &
    \texttt{en} &
    \faMicrophone\;\:\:Political debates (US pres.~campaigns) &
    span &
    6\,\, &
    1,989\,\, &
    \xmark &
    \xmark \\

    RuFal~\citep{shultz-2024-entity} &
    \texttt{en} &
    \faRss\;\;Political discourse (Russian govt.) & 
    text &
    13\,\, &
    700\,\, &
    \xmark &
    \xmark \\
    
    \midrule
    \textbf{\name~(\emph{ours})} &
    \texttt{it} &
    \faRss\;\;Climate change;\,migration;\,public health & 
    span &
    20\,\, &
    11,064\,\, &
    \cmark &
    \cmark \\
    
    \bottomrule
  \end{tabular}
  }%
  \caption{\label{tab:previous-work} Existing fallacy detection datasets. \textbf{Genres}. \faNewspaper[regular]~news;~\faBookReader~educational;~\faRss~social media/forums; \faMicrophone~transcripts. \textbf{Scope} of annotation. \emph{pair}:~question-answer/post-reply pair; \emph{text}:~whole text (e.g., post, paragraph); \emph{snippet}:~text excerpt; \emph{span}:~sequence of tokens in text. \textbf{\#C}.~Number of fallacy classes. \textbf{\#I}.~Number of \emph{human}-annotated instances. \textbf{Multi}. Whether multiple/overlapping annotations for the same pair/text/snippet/span are provided. \textbf{HLV}. Whether the dataset includes human label variation. $^\dagger$:~it also includes a negative class whose examples are not counted in \#I for fair comparison. $^\ddagger$:~it includes 12K extra examples but are \emph{machine}-annotated. *:~Arbitrary dataset name.}
\end{table*}

Fallacy detection is an open challenge in NLP and has shown to be intrinsically difficult for both humans and machines~\citep{alhindi-etal-2022-multitask}. Although some fallacy detection datasets have been proposed in recent years, they either contain coarse-grained annotations~\citep[e.g., post-level;][\emph{inter alia}]{jin-etal-2022-logical,habernal-etal-2018-name} or assume that no more than one fallacy can be expressed in a given text segment~\citep{sahai-etal-2021-breaking,goffredo-etal-2023-argument}. However, multiple fallacies may overlap in text~\citep{jin-etal-2022-logical} and knowing \emph{where} a fallacy occurs is central for educational purposes. Moreover, current datasets encode a single ``ground truth'' for fallacies through label aggregation. 
This cancels out human label variation~\citep{plank-2022-problem} that naturally occurs in fallacy annotation due to multiple plausible answers and genuine disagreement, in turn affecting modeling and evaluation.

We introduce \faina~\name~(\emph{\textbf{Fa}llacy detection with \textbf{in}dividual \textbf{a}nnotations}), a dataset for fallacy detection with human label variation (Figure~\ref{fig:overview}), along with experiments across setups of increasing complexity and thorough data analyses.

The \name~dataset is the first annotated resource for fallacy detection embracing multiple plausible answers and natural disagreement. It is also the first dataset providing annotations at the fine-grained level of text segments with potential overlaps, accounting for over 11K annotated fallacies by two expert annotators across an inventory of 20 fallacy types. \name~covers public discourse on migration, climate change, and public health issues in social media posts from Twitter, and focuses on Italian, a currently overlooked language in fallacy detection. 
To account for the complexity axes of the task, we design an evaluation framework that embraces the presence of multiple gold standards that are equally valid, partial span matches with overlaps, and the varying severity of fallacy classification errors.

We conduct experiments across four setups of increasing complexity, from post-level to span-level fallacy detection and using either fallacy macro-categories or the full inventory. 
Our results and analyses show that multi-task and multi-label transformer-based classifiers are strong baselines for fallacy detection tasks and that current large language models (LLMs) in zero-shot settings are still far from achieving satisfactory performance. 

We also provide thorough data analyses, including insights on our multi-round annotation procedure which minimizes annotation errors whilst keeping label variation, and a manual audit of the outputs generated by LLMs. To foster research on fine-grained fallacy detection and broadly on human label variation in span-level tasks, we make our materials available to the NLP community.


\section{Related Work} \label{sec:related-work}

The increasing interest in fallacy detection has led to the creation of datasets with different characteristics in recent years (Table~\ref{tab:previous-work}). 
Some works focused on the educational domain, either by collecting data through gamification~\citep{habernal-etal-2017-argotario,habernal-etal-2018-adapting} or from online quizzes~\citep{jin-etal-2022-logical}, whereas others studied transcripts of political debates~\citep{goffredo2022fallacious,goffredo-etal-2023-argument}. Fallacies in news articles have been explored by~\citet{jin-etal-2022-logical} and~\citet{alhindi-etal-2022-multitask} in climate change discourse, and by~\citet{musi2022developing} for analyzing COVID-19-related misinformation. 
Some works examined social media content using Reddit comments~\citep{habernal-etal-2018-name,sahai-etal-2021-breaking} or tweets~\citep{sheng-etal-2021-nice}, with a special interest on political discourse~\citep{macagno2022argumentation,shultz-2024-entity}.
We also analyze fallacies on social media but tackle previously unexplored topics over a four-year time frame and deal with the Italian language. 

Regarding annotation, current datasets mainly provide coarse-grained labels at the text (post, paragraph), snippet, or text pair level. The only exceptions are span-level datasets by~\citet{sahai-etal-2021-breaking} and~\citet{goffredo-etal-2023-argument} which, however, do not foresee overlapping annotations. Inspired by work on propaganda detection~\citep{da-san-martino-etal-2019-findings,da-san-martino-etal-2019-fine,da-san-martino-etal-2020-semeval} and persuasion techniques detection~\citep{piskorski-etal-2023-multilingual}, our dataset instead provides span-level annotations with overlaps for a better model transparency. Previous datasets include up to 3.6K \emph{human}-annotated instances~\citep{habernal-etal-2018-name} across 14 fallacies~\citep{goffredo2022fallacious}. Instead, ours provides 11K human-labeled spans across 20 fallacy types.

Recent work highlighted the importance of considering human label variation~\citep{plank-2022-problem}, i.e., genuine disagreement~\citep{poesio-artstein-2005-reliability}, subjectivity and perspectives~\citep{aroyo-and-welty-2015-truth,cabitza-etal-2023-toward}, and multiple plausible answers~\citep{nie-etal-2020-learn} as signal rather than noise. Yet, in fallacy detection all the labels from multiple annotators are typically aggregated~\citep{sahai-etal-2021-breaking}, selectively chosen~\citep{musi2022developing,habernal-etal-2017-argotario}, or adjudicated through discussion or by an expert~\citep[][\emph{inter alia}]{jin-etal-2022-logical,macagno2022argumentation,goffredo2022fallacious,goffredo-etal-2023-argument}. 
More broadly, label variation has been mostly studied at the post or token level, with only few exceptions analyzing span-level disagreement in argument annotation~\cite{lindahl-2024-disagreement,hautli-janisz-etal-2022-disagreement}. 
In our work, we resolve annotation errors while keeping genuine disagreement and multiple plausible answers, propose the first fallacy detection dataset at the span level with parallel annotations, and use human label variation during evaluation.


\section{Harmonization of Fallacies} \label{sec:fallacies-overview}

Given that different sets of fallacies have been proposed in the literature, we selected our tagset by reviewing fallacy types from previous work (Section~\ref{sec:related-work}) and harmonizing them. 
We thus compiled a list of 41 fallacies and conducted a pilot annotation on 15\% of our data (Section~\ref{sec:data}). 
This exploratory phase allowed us to homogenize fallacy names and unify those with similar definitions under the same label (e.g., \{\emph{Post hoc}, \emph{False cause}\} $\rightarrow$ \emph{Causal oversimplification}), leading to a total of 20 fallacy types. 
Most fallacy definitions are derived from~\citet{musi2022developing} but also from~\citet{da-san-martino-etal-2019-fine}, reflecting the persuasive nature of posts in our data. Fallacy definitions are provided below (extended definitions and examples are in Appendix~\ref{app:annotation-guidelines-specific}):

\begin{enumerate}
    \item \textbf{Ad hominem} (\textsc{ah}): an excessive attack on an individual or a group;
    \item \textbf{Appeal to authority} (\textsc{aa}): appealing to an authority figure or a group consensus to support a thesis;
    \item \textbf{Appeal to emotion} (\textsc{ae}): manipulation of the recipient's emotions in order to win an argument;
    \item \textbf{Causal oversimplification} (\textsc{co}): the attributed causal relation is simplified and fallacious;
    \item \textbf{Cherry picking} (\textsc{cp}): choosing evidence which supports a given position, dismissing findings which do not;
    \item \textbf{Circular reasoning} (\textsc{cr}): the end of an argument comes back to the beginning without having proven itself;
    \item \textbf{Doubt} (\textsc{do}): questioning the credibility of someone or something;
    \item \textbf{Evading the burden of proof} (\textsc{ep}): a position is advanced without any support as if it was self-evident;
    \item \textbf{False analogy} (\textsc{fa}): two different things or situations are treated equally;
    \item \textbf{False dilemma} (\textsc{fd}): a claim presenting only two options or sides when there are many;
    \item \textbf{Flag waving} (\textsc{fw}): playing on strong national feeling (or with respect to a group) to justify or promote an action or idea;
    \item \textbf{Hasty generalization} (\textsc{hg}): a generalization is drawn from a sample which is not representative or not applicable to the whole situation;
    \item \textbf{Loaded language} (\textsc{ll}): using words/phrases with strong emotional implications (positive or negative) to influence an audience;
    \item \textbf{Name calling or labeling} (\textsc{nc}): labeling the object of the propaganda campaign as either something the audience fears, hates, finds undesirable or otherwise loves or praises;
    \item \textbf{Red herring} (\textsc{rh}): the argument supporting the claim diverges the attention to issues which are irrelevant for the claim at hand;
    \item \textbf{Slippery slope} (\textsc{ss}): implies that an improbable or exaggerated consequence could result from a particular action;
    \item \textbf{Slogan} (\textsc{sl}): a brief and striking phrase used to provoke excitement of the audience;
    \item \textbf{Strawman} (\textsc{st}): the arguer misinterprets an opponent’s argument for the purpose of more easily attacking it, demolishes it, and then concludes that the opponent’s real argument has been demolished;
    \item \textbf{Thought-terminating cliché} (\textsc{tc}): short and generic phrase that discourages meaningful discussion;
    \item \textbf{Vagueness} (\textsc{va}): words which are ambiguous are shifted in meaning in the process of arguing or are left vague, being potentially subject to skewed interpretations.
\end{enumerate}

We further organize these fallacy types into a taxonomy grouped around three macro-categories which include all the others: \emph{Insufficient proof}, \emph{Simplification}, and \emph{Distraction}, following similar efforts~\citep{dimitrov-etal-2024-semeval,tindale2007fallacies}. The full taxonomy is presented in Figure~\ref{fig:taxonomy} and serves for evaluation purposes (Section~\ref{sec:experiments}).


\section{\faina~\name~Dataset} \label{sec:data}

In this section, we describe the creation of \name, from data collection (Section~\ref{sec:data-collection}) to data annotation (Section~\ref{sec:data-annotation}). We then provide detailed dataset statistics (Section~\ref{sec:data-statistics}). Data statements~\citep{bender-friedman-2018-data} are available in Appendix~\ref{app:data-statements}.

\subsection{Data Collection and Sampling} \label{sec:data-collection}

We collect social media posts in Italian (\texttt{it}) that discuss issues pertaining to migration, climate change, and public health using the Twitter APIs.\footnote{Data was retrieved in February 2023 when the social media platform X was still named Twitter, and when APIs for research purposes were still available for free.} To minimize temporal and topic biases, the posts were collected from a large time frame of four years (from \texttt{2019-01-01} to \texttt{2022-12-31}). We filter messages on the aforementioned topics by using a manually curated list of 436 keywords derived from trustable glossaries and manuals and extended to cover all applicable grammatical genders and numbers (see Appendix~\ref{app:search-keywords} for keywords and sources). We then retain posts with $\geq5$ tokens\footnote{Indeed, texts with $<5$ tokens are unlikely to contain argumentation and thus fallacies. For tokenization we used the \texttt{it\_core\_news\_sm} \texttt{spaCy} model (v3.5; \url{https://spacy.io}).} and select those with the largest number of likes and retweets as in~\citet{nakov2022overview}, therefore focusing on the messages with highest impact to the society. Specifically, to simultaneously mitigate topic, author, and temporal bias in sampling, we keep the top-$k$ posts ($k=10$) for each month and topic, further excluding messages authored by the same user after their most impactful post, and resampling messages until we obtain $k$ posts for each month-topic combination.\footnote{We found rare cases in which the same message appeared across more topics. We kept the post from the subset with the highest ranking and resample a new post from other subset(s).} As a result, we collect 1,440 posts balanced across topics (480 per topic) and time (360 per year) for fine-grained, span-level annotation with overlaps. 

\subsection{Manual Data Annotation} \label{sec:data-annotation}

Span-level annotation with an inventory of 20 fallacy types and potential overlaps is an intrinsically difficult task. It requires annotators to devote significant effort in understanding the nuances of fallacies and master the annotation guidelines, both for span segmentation and labeling. We thus avoid crowdsourcing which, as also noted by~\citet{da-san-martino-etal-2019-fine} for span-level propaganda annotation, is not suitable in this context. We instead devise an annotation protocol that foresees multiple rounds of annotation and discussion among two expert annotators ($\mathcal{A}_1$ and $\mathcal{A}_2$) which allows us to minimize annotation errors whilst keeping signals of human label variation. Details on annotators' profiles are in Appendix~\ref{app:data-statements}.
The annotation proved very challenging and took about 380 person-hours to be completed, discussions included.

\paragraph{Annotation protocol and guidelines} After a pilot phase in which we formalize the annotation guidelines and consolidate the label set (Section~\ref{sec:fallacies-overview}), we conduct five rounds of annotation and discussion in an increasingly larger number of posts (i.e., 60, 120, 180, 360, 720) balanced across topics. 
At each round, annotators \emph{i)} individually located all the text segments expressing fallacies, giving each of them one of the 20 labels from our inventory,\footnote{We use the INCEpTION annotation platform~\citep{klie-etal-2018-inception} since it supports span-level annotation with overlaps.} \emph{ii)} discussed with each other the annotated instances that diverged in the span extent and/or the assigned label, and \emph{iii)} resolved the cases of disagreement due to errors or attention drops, therefore keeping the naturally-occurring variation in annotation due to multiple plausible answers (e.g., interpretation) and genuine disagreement~\citep{plank-2022-problem}. The guidelines and annotations from previous rounds have been updated at the end of each round based on the discussion outcomes. After completing all the rounds, the two annotators also revised their annotations to ensure that all posts were labeled consistently. We provide the annotation guidelines in Appendix~\ref{app:annotation-guidelines} to foster future work covering other languages and additional annotators.

\begin{figure}[!t]
    \resizebox{\linewidth}{!}{%

    \begin{minipage}[t]{.625\linewidth}
    \centering
    \strut\vspace*{-9mm}\newline
    
        \begin{tikzpicture}
            \begin{axis}[
                xlabel={round},
                ylabel={IAA ($\gamma$)},
                xmin=0.8, xmax=5.2,
                ymin=0.2, ymax=0.8,
                xtick={1,2,3,4,5},
                ytick={0.8,0.7,0.6,0.5,0.4,0.3,0.2},
                height=5cm,
                width=4.5cm,
                legend pos=south west,
                ymajorgrids=true,
                grid style=dashed,
                title style={yshift=-3mm},
                legend style={yshift=-20mm,xshift=20mm},
                legend columns=-1,
                title={Span identification}
            ]

            \addplot[
                color=red!75!white,
                mark=*,
                ]
                coordinates {
                (1,0.3857821393988512)(2,0.4371931547450958)(3,0.4686853509164328)(4,0.49652522783298525)(5,0.4759202686989723)
                };
            \addlegendentry{before}

            \addplot[
                color=green!60!black,
                mark=*,
                ]
                coordinates {
                (1,0.6487089802126298)(2,0.6860159984821426)(3,0.6233792631234466)(4,0.6321975576824279)
                (5,0.6079000891877535)};
            \addlegendentry{after}
            \end{axis}
        \end{tikzpicture}
        

    \end{minipage}\hfill
    \begin{minipage}[t]{.625\linewidth}
    \centering
    \strut\vspace*{-9mm}\newline
        
        \begin{tikzpicture}
            \begin{axis}[
                xlabel={round},
                ylabel={IAA ($\gamma_{cat}$)},
                xmin=0.8, xmax=5.2,
                ymin=0.2, ymax=0.8,
                xtick={1,2,3,4,5},
                ytick={0.8,0.7,0.6,0.5,0.4,0.3,0.2},
                height=5cm,
                width=4.5cm,
                legend pos=south west,
                ymajorgrids=true,
                grid style=dashed,
                title style={yshift=-3mm},
                legend style={yshift=-20mm, draw=none},
                title={Span classification}
            ]

            \addplot[
                color=red!75!white,
                mark=*,
                ]
                coordinates {
                (1,0.2784156236556151)(2,0.3280955837910706)(3,0.3677531504002608)(4,0.40771429851679253)(5,0.40363390145288314)
                };
            
            \addplot[
                color=green!60!black,
                mark=*,
                ]
                coordinates {
                (1,0.5478709892625211)(2,0.6026936344138025)(3,0.5336831054261926)(4,0.5462384763261199)
                (5,0.531095305115031)};
            \end{axis}
        \end{tikzpicture}
    
    \end{minipage}
}%
	\caption{Inter-annotator agreement (IAA) scores for both span identification ($\gamma$) and classification ($\gamma_{cat}$) at each annotation round, \emph{before} and \emph{after} discussion.}
	\label{fig:iaa}
\end{figure}
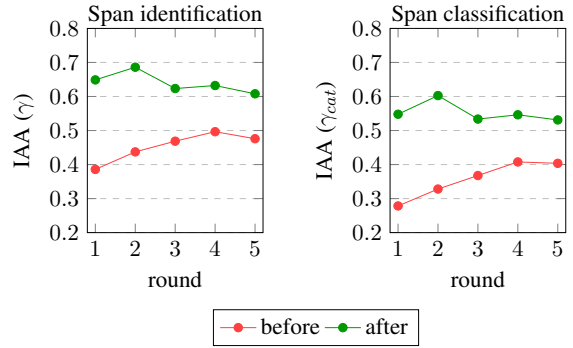

\paragraph{Inter-annotator agreement} We calculate the inter-annotator agreement (IAA) at each annotation round, both \emph{before} and \emph{after} discussion, and considering either span identification or span classification. This allows us to get insights on the whole annotation process besides final results only. Since our dataset includes span annotations of varying length which may also overlap, we avoid using Krippendorff's alpha~\citep[$\alpha$;][]{hayes-and-krippendorff-2007-answering} or other metrics that are typically applied at the token-level. We instead use $\gamma$~\citep{mathet-etal-2015-unified-v2} and $\gamma_{cat}$~\citep{mathet-2017-agreement-v2} as implemented in the \texttt{pygamma-agreement} (v0.5.9) package~\citep{titeux-and-riad-2021-pygamma} to better account for span-level identification and classification with overlaps. 
The IAA on the final dataset is $\gamma = 0.6240$ (\emph{span identification}) and $\gamma_{cat} = 0.5445$ (\emph{span classification}), which is comparable to results reported on the related span-level propaganda detection task~\citep{da-san-martino-etal-2019-fine}. 
By taking a closer look at the IAA over rounds (Figure~\ref{fig:iaa}), we notice that the difference between scores \emph{before} and \emph{after} discussion becomes less pronounced over rounds. This indicates that annotators increasingly align to each other, but also that annotating fallacies at the span level inherently calls for multiple discussions. At the same time, we notice that $\gamma$ and $\gamma_{cat}$ upper bounds (Figure~\ref{fig:iaa}; \emph{green line}) are rather stable over rounds (i.e., $[0.6, 0.7]$ and $[0.5, 0.6]$ for $\gamma$ and $\gamma_{cat}$, respectively), confirming that human label variation can be resolved only partially. 
Overall, identifying fallacy spans appears to be the main bottleneck for IAA, since finding and assigning a label to a text segment decreases the IAA only minimally (cf.~Figure~\ref{fig:iaa}; \emph{left} vs \emph{right}). Finally, we can see that annotators are more resistant to find a consensus after the initial rounds (cf.~Figure~\ref{fig:iaa}; \emph{fourth} vs \emph{fifth} round). We hypothesize that this is due to their increasing understanding of fallacies' nuances. We report the IAA across fallacies in Table~\ref{tab:stats-and-iaa}, in which \emph{Doubt} and \emph{Slogan} emerge as the fallacies with highest IAA whereas \emph{Cherry picking} and \emph{Vagueness} are the hardest ones to agree on.

\subsection{Data Statistics and Analysis} \label{sec:data-statistics}

Overall, \name~consists of 11,064 annotated spans (5,532$_{\pm253}$ per annotator) across 58,490 tokens in 1,440 social media posts. 
We present per-fallacy statistics on all annotations in Table~\ref{tab:stats-and-iaa} and refer to Appendix~\ref{sec:additional-dataset-details} for individual statistics for $\mathcal{A}_1$ and $\mathcal{A}_2$. 

\begin{table}[!t]
  \centering
  \resizebox{1.0\linewidth}{!}{%
  \begin{tabular}{rlrr|a}
    \toprule
     \textbf{Id} & \textbf{Fallacy type} & \textbf{Spans} & \textbf{Length} & \multicolumn{1}{r}{\textbf{$\gamma_{cat}$}} \\
    \midrule
    \textsc{ah} & Ad hominem & 319 & 16.0$_{\pm 13.3}$ & 0.6651 \\
    \textsc{aa} & Appeal to authority & 213 & 6.4$_{\; \: \pm 4.3}$ & 0.6147 \\
    \textsc{ae} & Appeal to emotion & 2,049 & 5.1$_{\; \: \pm 4.9}$ & 0.4730 \\
    \textsc{co} & Causal oversimplification & 142 & 19.0$_{\pm 10.7}$ & 0.5282 \\
    \textsc{cp} & Cherry picking & 94 & 28.8$_{\pm 12.3}$ & 0.3415 \\
    \textsc{cr} & Circular reasoning & 20 & 26.8$_{\pm 11.0}$ & 0.5397 \\
    \textsc{do} & Doubt & 482 & 16.1$_{\pm 11.4}$ & 0.7103 \\
    \textsc{ep} & Evading the burden of proof & 406 & 16.2$_{\; \: \pm 9.9}$ & 0.4335 \\
    \textsc{fa} & False analogy & 239 & 22.1$_{\pm 13.4}$ & 0.5243 \\
    \textsc{fd} & False dilemma & 90 & 15.9$_{\pm 11.0}$ & 0.5568 \\
    \textsc{fw} & Flag waving & 393 & 4.3$_{\; \: \pm 4.9}$ & 0.5735 \\
    \textsc{hg} & Hasty generalization & 464 & 11.2$_{\; \: \pm 8.0}$ & 0.4980 \\
    \textsc{ll} & Loaded language & 2,484 & 2.5$_{\; \: \pm 2.7}$ & 0.4365 \\
    \textsc{nc} & Name calling or labeling & 1,124 & 2.6$_{\; \: \pm 1.7}$ & 0.5566 \\
    \textsc{rh} & Red herring & 257 & 13.0$_{\; \: \pm 8.5}$ & 0.4378 \\
    \textsc{ss} & Slippery slope & 172 & 10.8$_{\; \: \pm 6.8}$ & 0.6552 \\
    \textsc{sl} & Slogan & 384 & 3.5$_{\; \: \pm 3.1}$ & 0.7101 \\
    \textsc{st} & Strawman & 109 & 36.3$_{\pm 15.4}$ & 0.5570 \\
    \textsc{tc} & Thought-terminating cliché & 285 & 5.2$_{\; \: \pm3.0}$ & 0.5305 \\
    \textsc{va} & Vagueness & 1,338 & 9.1$_{\; \: \pm 8.6}$ & 0.3701 \\
    \midrule
    & \textbf{All} & 11,064 & 7.6$_{\; \: \pm 9.3}$ & 0.5445 \\
    \bottomrule
  \end{tabular}
  }%
  \caption{\label{tab:stats-and-iaa} Statistics and per-class IAA scores ($\gamma_{cat}$) across all fallacy types. We report the number of \emph{spans} and their average \emph{length} (with standard deviation) at the token level considering all annotations. Individual statistics for both $\mathcal{A}_1$ and $\mathcal{A}_2$ are provided in~Appendix~\ref{sec:additional-dataset-details}.}
\end{table}

Fallacy spans have a length of 7.6$_{\pm9.3}$ tokens on average, but length greatly varies across fallacy types. The shortest fallacies are those related to language use (\emph{language fallacies} in~\citet{tindale2007fallacies}) such as \emph{Loaded language}, \emph{Name calling or labeling}, and \emph{Slogan} ($<4$ tokens), whereas the longest ones are those commonly referred to as \emph{logical fallacies} and \emph{fallacies of diversion}~\citep{tindale2007fallacies}, such as \emph{Strawman}, \emph{Cherry picking}, and \emph{Circular reasoning} ($>25$ tokens). 
The most and least frequent fallacies are \emph{Loaded language} (1,242$_{\pm178}$) and \emph{Circular reasoning} (10$_{\pm2}$), respectively. The number of annotated spans and average token length for each fallacy type varies between annotators. For instance, \emph{False analogy} has been annotated more by $\mathcal{A}_1$ than by $\mathcal{A}_2$ (147 \emph{vs} 92 spans), but on average $\mathcal{A}_2$ labeled longer text segments than $\mathcal{A}_1$ for that fallacy type, also with a higher standard deviation (24.1$_{\pm13.8}$ \emph{vs} 20.8$_{\pm13.0}$ tokens) (Appendix~\ref{sec:additional-dataset-details}).

\begin{figure}[t]
  \includegraphics[width=\columnwidth]{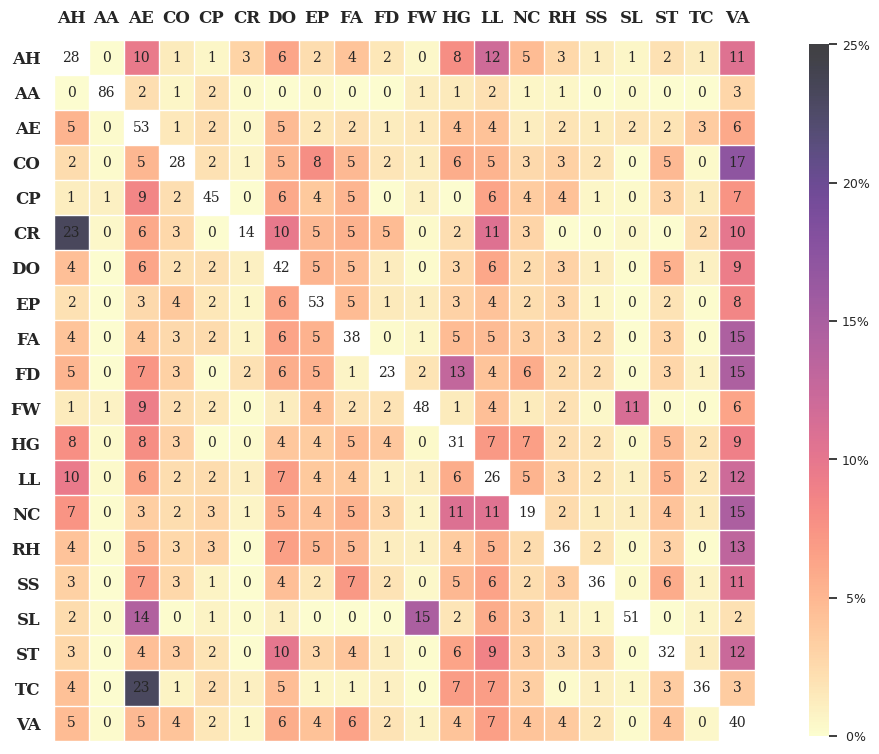}
  \caption{Overlap of fallacy annotations in terms of token percentages. Each row indicates the percentage of tokens for a given fallacy type that overlaps with any other fallacy type (\emph{columns}). White cells (\emph{diagonal}) indicate the percentage of tokens for each fallacy type that does not overlap with any other fallacy type. Fallacy overlaps for $\mathcal{A}_1$ and $\mathcal{A}_2$ annotations are in Appendix~\ref{sec:additional-dataset-details}.}
  \label{fig:overlaps-full}
\end{figure}

The \name~dataset has dense annotation (3.8$_{\pm0.2}$ spans/post) and overlaps among fallacy spans are also very frequent. In Figure~\ref{fig:overlaps-full} we show the percentage of pairwise overlaps at the token level across fallacy types considering all annotations (individual figures for $\mathcal{A}_1$ and $\mathcal{A}_2$ are in Appendix~\ref{sec:additional-dataset-details}). The percentage of tokens without overlaps is also shown (Figure~\ref{fig:overlaps-full}; \emph{diagonal cells}): for all fallacy types except \textsc{aa}, \textsc{ae}, \textsc{ep}, and \textsc{sl}, at least half of the tokens co-occur with other fallacies' tokens. 
Among the most frequent overlaps are the fallacies \emph{Thought-terminating cliché} and \emph{Appeal to emotion} (23\%$_{\pm2}$), since emotional words are often used in thought-stopping discussions.

\section{Experiments} \label{sec:experiments}

In this section, we present experiments on fallacy detection using the \name~dataset. We first introduce our experimental setup (Section~\ref{sec:experimental-setup}) and the approaches we employed (Section~\ref{sec:models}). We then provide results and a thorough discussion, including insights for future work (Section~\ref{sec:results-and-discussion}).

\subsection{Experimental Setup} \label{sec:experimental-setup}

\begin{figure*}[!ht]
  \includegraphics[width=\textwidth]{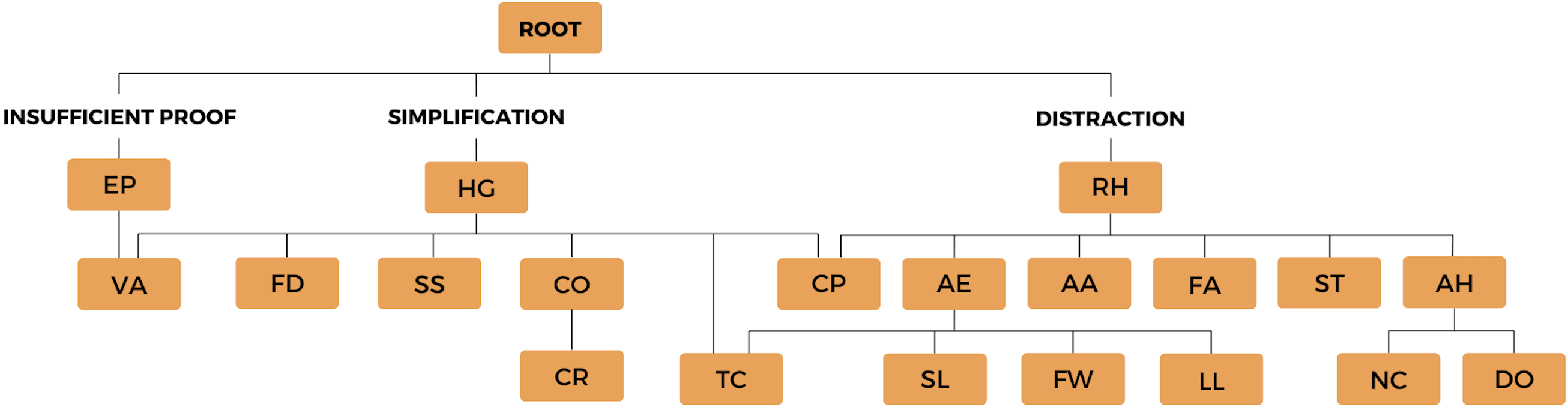}
  \caption{Taxonomy of fallacy types that we designed for evaluation purposes. Labels below the root (i.e., \emph{Insufficient proof}, \emph{Simplification}, and \emph{Distraction}) represent the fallacy macro-categories for \textsc{post-c} and \textsc{span-c} tasks.}
  \label{fig:taxonomy}
\end{figure*}

\paragraph{Tasks}
We cast fallacy detection into different tasks along two dimensions: the annotation unit, i.e., post-level (\textsc{post}) \emph{vs} span-level (\textsc{span}), and the classification granularity, i.e., coarse-grained with 3 fallacy macro-categories (\textsc{c}; Section~\ref{sec:fallacies-overview})\footnote{For \textsc{post} tasks, we assign to the post the set of unique fallacy span types occurring in the message. For coarse-grained setups, we map each fallacy type to its corresponding macro-categories according to our taxonomy (Figure~\ref{fig:taxonomy}).} \emph{vs} fine-grained with all the 20 fallacy types (\textsc{f}). As a result, we deal with four subtasks of increasing complexity (i.e., \textsc{post-c}, \textsc{post-f}, \textsc{span-c}, and \textsc{span-f}) implying the use of different evaluation metrics as detailed in the following.

\paragraph{Data splits} For the sake of the experiments, we divide \name~into $k$ \texttt{train} and \texttt{test} data splits using $k$-fold cross validation ($k=5$). We use the \texttt{train} data portions for pretrained models to fine-tune, and use the \texttt{test} portion for evaluation. For model selection, we rely on the \texttt{train} splits, further diving them into \texttt{train} (80\%) and \texttt{dev} (20\%), and selecting the best model configuration based on average performance on \texttt{dev} data portions. Each data split contains annotations for the same posts, at either the post or span level, by both  annotators.

\paragraph{Evaluation metrics}

We evaluate performance using different flavors of precision, recall, and F$_1$ score to capture the diverse challenges of each task. 
For \textsc{post} tasks, in which the set of fallacies expressed in a post must be predicted, we use standard micro-averaged scores. 
For the more challenging \textsc{span} tasks, in which all text segments expressing fallacies within a post must be identified and classified, we instead adopt precision, recall, and F$_1$ variants proposed by~\citet{da-san-martino-etal-2019-fine} and extend them to operate on tokens.\footnote{The metrics in~\citet{da-san-martino-etal-2019-fine} originally work on characters for determining partial span matches. However, we observe that operating on tokens makes the metric less dependant on the length of the tokens themselves.} We use such variants since they are expressly designed for tasks with spans of varying length that may overlap. 
Moreover, to account for the severity of labeling errors (e.g., predicting \emph{Red herring} instead of \emph{Appeal to authority} is less problematic than predicting \emph{False dilemma}), we compute results using both \emph{strict} and \emph{soft} evaluation modes in the \textsc{span-f} task. Concretely, while in strict mode we reward models only if they predict the intended label, in soft mode we give them partial credit if the predicted label is an immediate parent of the actual label (Figure~\ref{fig:taxonomy}). This is achieved by setting $\delta=0.5$ as coefficient for partial label matches in the distance function of the metric by~\citet{da-san-martino-etal-2019-fine}.

\paragraph{Multiple gold standards} 
\name~consists of multiple parallel annotations (i.e., \emph{multiple views}) that are equally reliable. To equally account for all test set versions while avoiding to favor those with more/less annotations (and thus avoiding to favor models that over/under-predict fallacies), we macro-average scores on individual test sets. The simplicity of this approach makes it suitable for extending evaluation on future test set versions.

\subsection{Models} \label{sec:models}

We perform experiments on all tasks using different models in supervised and unsupervised settings. 

\paragraph{Supervised models} Since each data instance in \name~has multiple ``ground truths'', i.e., parallel labels by annotators $\mathcal{A}_1, \mathcal{A}_2 \in \mathcal{A}$, in the supervised setting we aim to jointly model all $|\mathcal{A}|$ annotation versions (hereafter, \emph{views}) to account for human label variation. We therefore adopt a multi-task learning~\citep{caruana1997multitask} approach to leverage signals of different but related \emph{views} through a shared encoder and $D$ decoders. 
Specifically, for \textsc{post} tasks we propose a \emph{multi-view, multi-label} (MVML) model that uses $D=|\mathcal{A}|$ decoders and thus outputs $D$ sets of predicted labels, one for each \emph{view} containing all labels that exceed a threshold $\tau$.
For \textsc{span} tasks, which are sequence labeling problems relying on the \textsc{bio}-tagging scheme, we instead propose a \emph{multi-view, multi-decoder} (MVMD) model, with a separate decoder for each \emph{view} and fallacy type $f \in \mathcal{F}$ (i.e., $D = |\mathcal{A} \times \mathcal{F}|$). 
We give equal importance to all decoders, i.e., computing the multi-task learning loss as $L = \sum_d \lambda_d L_d$, where $L_d$ is the loss for task $d \in D$ and $\lambda_d=1$ the corresponding weight.
For all tasks, we use widespread models pretrained on Italian data as encoders, namely AlBERTo~\citep{polignano-etal-2019-alberto} and UmBERTo~\citep{parisi-etal-2020-umberto}, leading to \textsc{MVML-alb} and \textsc{MVML-umb} models for \textsc{post} tasks, and \textsc{MVMD-alb} and \textsc{MVMD-umb} for \textsc{span} tasks. We implement our models using the MaChAmp v0.4 toolkit~\citep{van-der-goot-etal-2021-massive} and adopt default hyper-parameter values (Appendix~\ref{app:additional-experimental-details}).

\paragraph{Unsupervised models}
Given the challenging nature of our tasks, we further assess classification performance with instruction-tuned LLMs in a zero-shot setting. We experiment with LLaMa-3 8B~\citep{dubey2024llama} and Mixtral 8x7B~\citep{jiang-etal-2024-mixtral} to favor reproducibility, as they are freely available and widely used models. Moreover, both include Italian in their pretraining data. We design prompts that describe each task and output format, and either include fallacy definitions (Section~\ref{sec:fallacies-overview}) or just fallacy names (see Appendix~\ref{app:additional-experimental-details}).\footnote{We tested our prompts in both English and Italian. However, our preliminary experiments showed that prompts in Italian led to inconsistent outputs with regards to fallacy names. We thus employ prompts in English to favor I/O consistency.} 
During model selection, we observe that including definitions increased performance on the \texttt{dev} splits. Therefore, we select \emph{zero-shot} models \emph{with definitions} (ZSWD) for testing, i.e., \textsc{ZSWD-llama} and \textsc{ZSWD-mixtr}. Being unsupervised, these models naturally yield a single output for each data instance. We thus compare this output against all $|\mathcal{A}|$ data instance \emph{views} during evaluation. We use the Hugging Face library 
and employ default model hyperparameters.

\subsection{Results and Discussion} \label{sec:results-and-discussion}

Results across all task setups are reported in Table~\ref{tab:results} (individual results are in Appendix~\ref{app:additional-experimental-details}). 
We observe that our \emph{multi-view} approaches (i.e., \textsc{MVML-*} and \textsc{MVMD-*}) yield promising results both in post- and span-level tasks, offering well-performing baselines upon which future approaches can be built. Concerning the encoders, UmBERTo (i.e., \textsc{*-umb} models) appears unstable on fine-grained tasks, while it achieves similar performance to AlBERTo (i.e., \textsc{*-alb} models) in coarse-grained classification. The difference in performance between the two may be due to the former being pretrained on the Italian portion of the OSCAR corpus~\citep{suarez2019asynchronous}, while AlBERTo is trained exclusively on tweets, and thus is more in line with \name~data. 
As expected, classification at the span level with fine-grained labels (i.e., \textsc{span-f}) is the hardest task setup, although the \textsc{MVMD-alb} model achieves a F$_1$ score of 33.3 and 37.0 using \emph{strict} and \emph{soft} evaluation modes, respectively. These results are in line with performance on span-level tasks of similar complexity~\citep{da-san-martino-etal-2019-fine} and can be seen as strong baselines considering that 20 fallacy types are involved. We also observe that high recall is more challenging than high precision, probably because of the presence of multiple labels across partially and fully-overlapping spans.

As regards the second set of experiments using generative LLMs in a zero-shot setting (i.e, \textsc{ZSWD-*} models), results appear unreliable. While we acknowledge that fine-tuning approaches cannot be directly compared with zero-shot classification, our main goal was instead to assess to what extent we can expect to challenge traditional supervised approaches with zero-shot generative models.\footnote{We also tested LLMs in few-shot settings, but preliminary experiments showed that models tended to replicate the characteristics of the example(s) provided, such as span lengths and the number of fallacies, making the results untrustworthy. Future research is still needed to study and mitigate the brittleness of the few-shot approach in similar setups as ours.} Our results show that a complex task like fallacy detection, which involves capturing lexical, semantic, and even pragmatic aspects of communication, is still far from being addressed with generative models, especially if we aim at embracing human label variation. In addition to the low performance across the challenging \textsc{span} setups (e.g., F$_1$ of 3.4--5.0 and 4.2--6.5 on the \textsc{span-f} task for \textsc{ZSWD-llama} and \textsc{ZSWD-mixtr}, respectively), LLMs are also more computationally expensive than the other proposed models, making them impractical in scenarios in which computational time is a requirement or large computational resources are not available.

\begin{table}[t!]
  \centering
  \resizebox{1.0\linewidth}{!}{%
  \begin{tabular}{clrra}
    \toprule
    & \textbf{Model} & \textbf{P} & \textbf{R} & \multicolumn{1}{r}{\textbf{F$_1$}} \\
    \midrule
    \multirow{4}{*}{\rotatebox[origin=c]{90}{\textsc{\textbf{POST-C}}}} &
    \textsc{MVML-alb} & 80.0$_{\pm1.5}$ & 74.0$_{\pm2.3}$ & \textbf{76.8}$_{\pm1.6}$ \\
    & \textsc{MVML-umb} & 84.5$_{\pm1.3}$ & 70.1$_{\pm4.2}$ & 76.6$_{\pm2.8}$ \\
    & \textsc{ZSWD-llama} & 57.9$_{\pm1.9}$ & 70.0$_{\pm1.9}$ & 63.3$_{\pm1.5}$ \\
    & \textsc{ZSWD-mixtr} & 64.7$_{\pm1.6}$ & 45.2$_{\pm1.0}$ & 53.2$_{\pm1.2}$ \\
    \midrule
    \multirow{4}{*}{\rotatebox[origin=c]{90}{\textsc{\textbf{POST-F}}}} &
    \textsc{MVML-alb} & 63.0$_{\pm2.0}$ & 34.3$_{\pm1.9}$ & \textbf{44.3}$_{\pm1.9}$ \\
    & \textsc{MVML-umb} & 39.0$_{\pm3.7}$ & 14.6$_{\pm1.6}$ & 21.3$_{\pm2.2}$ \\
    & \textsc{ZSWD-llama} & 20.9$_{\pm1.5}$ & 24.3$_{\pm2.3}$ & 22.5$_{\pm1.8}$ \\
    & \textsc{ZSWD-mixtr} & 26.0$_{\pm1.8}$ & 18.1$_{\pm1.4}$ & 21.4$_{\pm1.5}$ \\
    \midrule
    \multirow{4}{*}{\rotatebox[origin=c]{90}{\textsc{\textbf{SPAN-C}}}} &
    \textsc{MVMD-alb} & 55.2$_{\pm1.7}$ & 51.7$_{\pm2.1}$ & 53.3$_{\pm1.4}$ \\
    & \textsc{MVMD-umb} & 59.8$_{\pm1.5}$ & 50.4$_{\pm2.4}$ & \textbf{54.7}$_{\pm1.5}$ \\
    & \textsc{ZSWD-llama} & 25.3$_{\pm4.2}$ & 7.0$_{\pm0.8}$ & 10.9$_{\pm0.9}$ \\
    & \textsc{ZSWD-mixtr} & 31.6$_{\pm1.2}$ & 20.9$_{\pm1.4}$ & 25.1$_{\pm1.2}$ \\
    \midrule
    \multirow{12}{*}{\rotatebox[origin=c]{90}{\textsc{\textbf{SPAN-F}}}} &
    \textbf{\emph{Strict mode}} &  &  & \\
    & ~~~\textsc{MVMD-alb} & 47.6$_{\pm1.9}$ & 25.6$_{\pm1.6}$ & \textbf{33.3}$_{\pm1.4}$ \\
    & ~~~\textsc{MVMD-umb} & 57.5$_{\pm5.9}$ & 3.9$_{\pm0.7}$ & 7.3$_{\pm1.3}$ \\
    & ~~~\textsc{ZSWD-llama} & 4.5$_{\pm0.5}$ & 2.7$_{\pm0.4}$ & 3.4$_{\pm0.3}$ \\
    & ~~~\textsc{ZSWD-mixtr} & 5.8$_{\pm1.1}$ & 3.2$_{\pm0.5}$ & 4.2$_{\pm0.7}$ \\
    & \textbf{\emph{Soft mode}} &  &  & \\
    & ~~~\textsc{MVMD-alb} & 52.2$_{\pm2.0}$ & 28.7$_{\pm1.7}$ & \textbf{37.0}$_{\pm1.5}$ \\
    & ~~~\textsc{MVMD-umb} & 66.3$_{\pm5.5}$ & 4.8$_{\pm0.7}$ & 8.9$_{\pm1.3}$ \\
    & ~~~\textsc{ZSWD-llama} & 6.4$_{\pm0.6}$ & 4.2$_{\pm0.5}$ & 5.0$_{\pm0.4}$ \\
    & ~~~\textsc{ZSWD-mixtr} & 8.2$_{\pm1.5}$ & 5.4$_{\pm1.0}$ & 6.5$_{\pm1.1}$ \\
    \bottomrule
    
  \end{tabular}
  }%
  \caption{\label{tab:results} Test set results for \textsc{post} and \textsc{span} tasks at the \emph{coarse-grained} (\textsc{c}) and \emph{fine-grained} (\textsc{f}) level. We report average precision (P), recall (R), and F$_1$ scores (w/ std dev) across $k=5$ splits, averaged over all $|\mathcal{A}|$ test versions. For \textsc{span-f}, we also present scores using both \emph{strict} and \emph{soft} modes. Best results are in bold. Results on individual test sets ($\mathcal{A}_1$ and $\mathcal{A}_2$) are in~Appendix~\ref{app:additional-experimental-details}.}
\end{table}

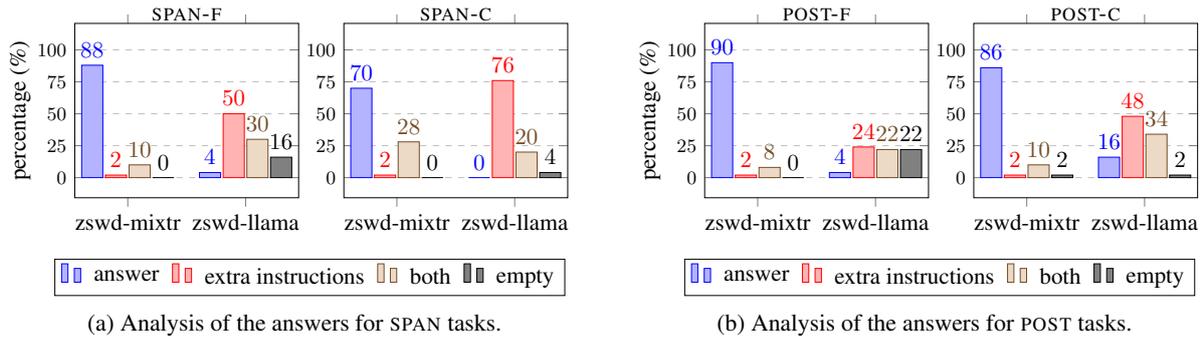
\begin{figure*}[!t]

\begin{subfigure}{.482\textwidth}

    \resizebox{\linewidth}{!}{%

\begin{minipage}[t]{.625\linewidth}
    \centering
    \strut\vspace*{-9mm}\newline

\begin{tikzpicture}
\begin{axis}[
    height=4.5cm,
    width=5.3cm,
    ymin=0, ymax=105,
    ybar=1.2pt,
    enlargelimits=0.15,
    enlarge x limits=0.45,
    legend style={at={(1.05,-0.35)},
      anchor=north, legend columns=-1, column sep=4pt},
    ylabel={percentage (\%)},
    ylabel style={yshift=-1.5mm},
    symbolic x coords={zswd-mixtr,zswd-llama},
    xtick=data,
    nodes near coords,
    nodes near coords align={vertical},
    title style={yshift=-3mm},
    title={\textsc{span-f}},
    ytick={0,25,50,75,100},
    ymajorgrids=true,
    grid style=dashed,
    yticklabel style = {font=\small}
]

\addplot coordinates {(zswd-mixtr,88) (zswd-llama,4)};
\addplot coordinates {(zswd-mixtr,2) (zswd-llama,50)};
\addplot coordinates {(zswd-mixtr,10) (zswd-llama,30)};
\addplot coordinates {(zswd-mixtr,0) (zswd-llama,16)};
\legend{answer, extra instructions, both, empty}
\end{axis}
\end{tikzpicture}

\end{minipage}\hfill
    \begin{minipage}[t]{.625\linewidth}
    \centering
    \strut\vspace*{-9mm}\newline

\begin{tikzpicture}
\begin{axis}[
    height=4.5cm,
    width=5.3cm,
    ymin=0, ymax=105,
    ybar=1.2pt,
    enlargelimits=0.15,
    enlarge x limits=0.45,
    legend style={at={(0.5,-0.35)},
      anchor=north, legend columns=-1, column sep=4pt},
    symbolic x coords={zswd-mixtr,zswd-llama},
    xtick=data,
    nodes near coords,
    nodes near coords align={vertical},
    title style={yshift=-3mm},
    title={\textsc{span-c}},
    ytick={0,25,50,75,100},
    ymajorgrids=true,
    grid style=dashed,
    legend style={draw=none},
    yticklabel style = {font=\small},
]

\addplot coordinates {(zswd-mixtr,70) (zswd-llama,0)};
\addplot coordinates {(zswd-mixtr,2) (zswd-llama,76)};
\addplot coordinates {(zswd-mixtr,28) (zswd-llama,20)};
\addplot coordinates {(zswd-mixtr,0) (zswd-llama,4)};
\legend{}
\end{axis}
\end{tikzpicture}

\end{minipage}

}%
	\caption{Analysis of the answers for \textsc{span} tasks.}
	\label{fig:generation-analysis-1}

\end{subfigure}%
\hfill
\begin{subfigure}{.482\textwidth}

    \resizebox{\linewidth}{!}{%

\begin{minipage}[t]{.625\linewidth}
    \centering
    \strut\vspace*{-9mm}\newline

\begin{tikzpicture}
\begin{axis}[
    height=4.5cm,
    width=5.3cm,
    ymin=0, ymax=105,
    ybar=1.2pt,
    enlargelimits=0.15,
    enlarge x limits=0.45,
    legend style={at={(1.05,-0.35)},
      anchor=north, legend columns=-1, column sep=4pt},
    ylabel={percentage (\%)},
    ylabel style={yshift=-1.5mm},
    symbolic x coords={zswd-mixtr,zswd-llama},
    xtick=data,
    nodes near coords,
    nodes near coords align={vertical},
    title style={yshift=-3mm},
    title={\textsc{post-f}},
    ytick={0,25,50,75,100},
    ymajorgrids=true,
    grid style=dashed,
    yticklabel style = {font=\small}
]

\addplot coordinates {(zswd-mixtr,90) (zswd-llama,4)};
\addplot coordinates {(zswd-mixtr,2) (zswd-llama,24)};
\addplot coordinates {(zswd-mixtr,8) (zswd-llama,22)};
\addplot coordinates {(zswd-mixtr,0) (zswd-llama,22)};
\legend{answer, extra instructions, both, empty}
\end{axis}
\end{tikzpicture}

\end{minipage}\hfill
    \begin{minipage}[t]{.625\linewidth}
    \centering
    \strut\vspace*{-9mm}\newline

\begin{tikzpicture}
\begin{axis}[
    height=4.5cm,
    width=5.3cm,
    ymin=0, ymax=105,
    ybar=1.2pt,
    enlargelimits=0.15,
    enlarge x limits=0.45,
    legend style={at={(0.5,-0.35)},
      anchor=north, legend columns=-1, column sep=4pt},
    symbolic x coords={zswd-mixtr,zswd-llama},
    xtick=data,
    nodes near coords,
    nodes near coords align={vertical},
    title style={yshift=-3mm},
    title={\textsc{post-c}},
    ytick={0,25,50,75,100},
    ymajorgrids=true,
    grid style=dashed,
    legend style={draw=none},
    yticklabel style = {font=\small}
]

\addplot coordinates {(zswd-mixtr,86) (zswd-llama,16)};
\addplot coordinates {(zswd-mixtr,2) (zswd-llama,48)};
\addplot coordinates {(zswd-mixtr,10) (zswd-llama,34)};
\addplot coordinates {(zswd-mixtr,2) (zswd-llama,2)};
\legend{}
\end{axis}
\end{tikzpicture}

\end{minipage}

}%
	\caption{Analysis of the answers for \textsc{post} tasks.}
	\label{fig:generation-analysis-3}

\end{subfigure}%

\caption{Analysis of the \textbf{raw outputs} generated by LLMs across the four tasks according to whether they contain an actual \emph{answer}, \emph{extra instructions}, \emph{both} an actual answer and extra instructions, or an \emph{empty} response.}
\label{fig:analysis-general}

\end{figure*}

\begin{figure*}[!t]

\begin{subfigure}{.482\textwidth}

    \resizebox{\linewidth}{!}{%

\begin{minipage}[t]{.625\linewidth}
    \centering
    \strut\vspace*{-9mm}\newline

\begin{tikzpicture}
\begin{axis}[
    height=4.5cm,
    width=5.3cm,
    ymin=0, ymax=105,
    ybar=1.2pt,
    enlargelimits=0.15,
    enlarge x limits=0.45,
    legend style={at={(1.03,-0.35)},
      anchor=north, legend columns=-1, column sep=4pt},
    ylabel={percentage (\%)},
    ylabel style={yshift=-1.5mm},
    symbolic x coords={zswd-mixtr,zswd-llama},
    xtick=data,
    nodes near coords,
    nodes near coords align={vertical},
    title style={yshift=-3mm},
    title={\textsc{span-f}},
    ytick={0,25,50,75,100},
    ymajorgrids=true,
    grid style=dashed,
    yticklabel style = {font=\small}
]

\addplot coordinates {(zswd-mixtr,92) (zswd-llama,28)};
\addplot coordinates {(zswd-mixtr,4) (zswd-llama,4)};
\addplot coordinates {(zswd-mixtr,2) (zswd-llama,2)};
\addplot coordinates {(zswd-mixtr,0) (zswd-llama,22)};
\legend{format ok, explain, wrong labels, repeat}
\end{axis}
\end{tikzpicture}

\end{minipage}\hfill
    \begin{minipage}[t]{.625\linewidth}
    \centering
    \strut\vspace*{-9mm}\newline

\begin{tikzpicture}
\begin{axis}[
    height=4.5cm,
    width=5.3cm,
    ymin=0, ymax=105,
    ybar=1.2pt,
    enlargelimits=0.15,
    enlarge x limits=0.45,
    legend style={at={(0.5,-0.35)},
      anchor=north, legend columns=-1, column sep=4pt},
    symbolic x coords={zswd-mixtr,zswd-llama},
    xtick=data,
    nodes near coords,
    nodes near coords align={vertical},
    title style={yshift=-3mm},
    title={\textsc{span-c}},
    ytick={0,25,50,75,100},
    ymajorgrids=true,
    grid style=dashed,
    legend style={draw=none},
    yticklabel style = {font=\small}
]

\addplot coordinates {(zswd-mixtr,94) (zswd-llama,18)};
\addplot coordinates {(zswd-mixtr,12) (zswd-llama,8)};
\addplot coordinates {(zswd-mixtr,8) (zswd-llama,0)};
\addplot coordinates {(zswd-mixtr,0) (zswd-llama,8)};
\legend{}
\end{axis}
\end{tikzpicture}

\end{minipage}

}%
	\caption{Analysis of the answers for \textsc{span} tasks.}
	\label{fig:generation-analysis-2}

\end{subfigure}%
\hfill
\begin{subfigure}{.482\textwidth}

    \resizebox{\linewidth}{!}{%

\begin{minipage}[t]{.625\linewidth}
    \centering
    \strut\vspace*{-9mm}\newline

\begin{tikzpicture}
\begin{axis}[
    height=4.5cm,
    width=5.3cm,
    ymin=0, ymax=105,
    ybar=1.2pt,
    enlargelimits=0.15,
    enlarge x limits=0.45,
    legend style={at={(1.03,-0.35)},
      anchor=north, legend columns=-1, column sep=4pt},
    ylabel={percentage (\%)},
    ylabel style={yshift=-1.5mm},
    symbolic x coords={zswd-mixtr,zswd-llama},
    xtick=data,
    nodes near coords,
    nodes near coords align={vertical},
    title style={yshift=-3mm},
    title={\textsc{post-f}},
    ytick={0,25,50,75,100},
    ymajorgrids=true,
    grid style=dashed,
    yticklabel style = {font=\small}
]

\addplot coordinates {(zswd-mixtr,94) (zswd-llama,24)};
\addplot coordinates {(zswd-mixtr,8) (zswd-llama,16)};
\addplot coordinates {(zswd-mixtr,2) (zswd-llama,0)};
\addplot coordinates {(zswd-mixtr,0) (zswd-llama,20)};
\legend{format ok, explain, wrong labels, repeat}
\end{axis}
\end{tikzpicture}

\end{minipage}\hfill
    \begin{minipage}[t]{.625\linewidth}
    \centering
    \strut\vspace*{-9mm}\newline

\begin{tikzpicture}
\begin{axis}[
    height=4.5cm,
    width=5.3cm,
    ymin=0, ymax=105,
    ybar=1.2pt,
    enlargelimits=0.15,
    enlarge x limits=0.45,
    legend style={at={(0.5,-0.35)},
      anchor=north, legend columns=-1, column sep=4pt},
    symbolic x coords={zswd-mixtr,zswd-llama},
    xtick=data,
    nodes near coords,
    nodes near coords align={vertical},
    title style={yshift=-3mm},
    title={\textsc{post-c}},
    ytick={0,25,50,75,100},
    ymajorgrids=true,
    grid style=dashed,
    legend style={draw=none},
    yticklabel style = {font=\small}
]

\addplot coordinates {(zswd-mixtr,94) (zswd-llama,46)};
\addplot coordinates {(zswd-mixtr,16) (zswd-llama,20)};
\addplot coordinates {(zswd-mixtr,0) (zswd-llama,0)};
\addplot coordinates {(zswd-mixtr,2) (zswd-llama,30)};
\legend{}
\end{axis}
\end{tikzpicture}

\end{minipage}

}%
	\caption{Analysis of the answers for \textsc{post} tasks.}
	\label{fig:generation-analysis-4}

\end{subfigure}%

\caption{Analysis of the \textbf{actual answers} (i.e., \emph{answer}+\emph{both}; Figure~\ref{fig:analysis-general}) generated by LLMs across the four tasks according to whether the output is in the requested format (\emph{format ok}), provides \emph{explain}ations, \emph{wrong labels}, or repetitions (\emph{repeat}). A single answer may meet more than one aspect, e.g., it can be both \emph{format ok} and \emph{repeat}.}
\label{fig:analysis-answers}

\end{figure*}
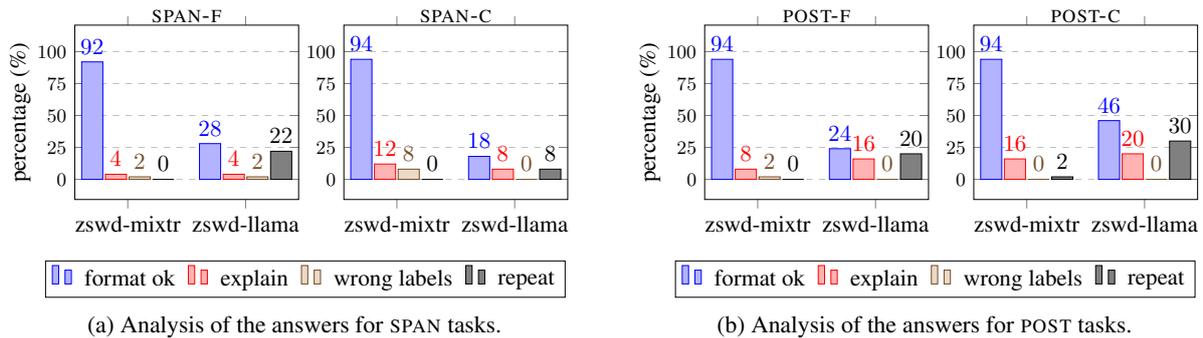

To further analyze the behavior of LLMs in fallacy detection, we conduct a manual analysis of the outputs of \textsc{ZSWD-llama} and \textsc{ZSWD-mixtr} across all task setups. We sample 50 outputs for each model and setup, for a total of 400 samples. We first audit them according to whether they provide an actual \emph{answer}, \emph{extra instructions}, \emph{both}, or an \emph{empty} response (Figure~\ref{fig:analysis-general}). Then, we analyze instances where an actual answer is provided, auditing if the output is in the requested format (\emph{format ok}), provides extra \emph{explain}ations, \emph{wrong labels}, or repetitions (\emph{repeat}) (Figure~\ref{fig:analysis-answers}).


For \textsc{span} tasks, \textsc{ZSWD-llama} provides few answers (20--34\%; \emph{answer}+\emph{both} in Figure~\ref{fig:generation-analysis-1}) compared to \textsc{ZSWD-mixtr} despite being instruction-tuned, and instead mainly generates prompt continuations (80--96\%; \emph{extra instructions}+\emph{both} in Figure~\ref{fig:generation-analysis-1}). \textsc{ZSWD-mixtr} appears more robust, producing up to 98\% answers (\emph{answer}+\emph{both} in Figure~\ref{fig:generation-analysis-1}), of which 92--94\% in the requested format (\emph{format ok} in Figure~\ref{fig:generation-analysis-2}). However, the results obtained with \textsc{ZSWD-mixtr} (Table~\ref{tab:results}) are just slightly higher than those of \textsc{ZSWD-llama}, suggesting that while \textsc{ZSWD-mixtr} produces answers in most cases, those are not actually reliable. 

As a comparison, we report the results for \textsc{post} tasks in Figure~\ref{fig:generation-analysis-3} and \ref{fig:generation-analysis-4}. We observe that the general trends highlighted for the \textsc{span} tasks still hold also for this setting. Overall, this qualitative analysis indicates that future work is needed for dealing with complex tasks such as fine-grained fallacy detection with LLMs in zero-shot setups.


\section{Conclusions} \label{sec:conclusions}
We introduced \name, the first fallacy detection dataset embracing multiple plausible answers and natural disagreement at the fine-grained level of text segments. 
\name~advances research on human label variation in NLP and opens new avenues for research. Given its multi-topic and multi-year nature, it can be used to benchmark fallacy detection approaches in out-of-domain scenarios and across time. Moreover, our annotation paradigm and guidelines can be applied to cover new languages, topics, and additional annotators.
Lastly, \name~can be used for novel work on cross-lingual annotation transfer in which source and target languages are shifted, e.g., from Italian to English.


\section*{Limitations}
Our dataset contains tweets in one language. Although the annotation scheme can be easily adopted for other languages, the findings and insights obtained from the experiments may not hold on other data sources and languages. 
For dataset creation and experiments with human label variation we rely on two annotators. We are aware that more perspectives could have been leveraged using crowd-sourcing platforms. However, we opted to involve only expert annotators to prioritize annotation reliability, further providing the full annotation guidelines and details on the annotation protocol to encourage future extensions.
The main focus of the paper is on the resource creation and assessment and its fine-grained evaluation; therefore, we employed a limited set of models in our experiments. The performance evaluation of additional models and the fine-tuning of LLMs is a research direction we would like to pursue in future work.


\section*{Ethics Statement}

The \name~dataset has been created by paying particular attention to data minimization principles and complying with privacy requirements. Data was collected using the Twitter APIs when they were still freely available for research purposes, and up to one tweet per user has been retrieved for each month and topic. It is therefore impossible to profile users, and this also avoids the users' writing style to interfere with classification performance. 

To follow good scientific practices, the \name~dataset is released including only the post texts, with no user information. Indeed, while in the past Twitter datasets were commonly released as a list of tweet IDs, this would make it hard to replicate our work because free Twitter APIs have been discontinued. The dataset is  released in its anonymized form, after replacing user mentions, email addresses, phone numbers and URLs with placeholders, as described in Appendix \ref{app:data-statements}.
To download \name, it is necessary to fill in an online form declaring compliance with user protection regulations and exclude data misuse. 


\section*{Acknowledgments}
This work was funded by the European Union's Horizon Europe research and innovation programme under grant agreement No.~101135437 (AI-CODE). 
We acknowledge also the support of the PNRR project FAIR -- Future AI Research (PE00000013), under the NRRP MUR program funded by NextGeneration EU.


\bibliography{custom}

\newpage


\newpage

\appendix

\section*{Appendix} \label{sec:appendix}

\section{Data Statements} \label{app:data-statements}

We present data statements~\citep{bender-friedman-2018-data} for \name~in the following.

\paragraph{\textsc{Curation rationale.}} 
The dataset consists of anonymized social media messages from Twitter with fallacy annotations at the span level. The posts were collected using search keywords related to migration, climate change, and public health whose use is generally not negatively/positively connoted to minimize the over-representation of specific stances on the topics (see Appendix~\ref{app:search-keywords}). The dataset was created to study fallacious argumentation in social media, to educate about critical thinking, and to encourage research on embracing human label variation in NLP. The dataset is in a CoNLL-like format with individual annotators' labels. Further details on data creation and annotation are in Section~\ref{sec:data}.

\paragraph{\textsc{Language varieties.}} 
The language represented in the dataset is Italian (\texttt{ita}) in the form of spontaneous written speech. Rare instances ($<1\%$) also exhibit code-switching between Italian and local language varieties of Italy~\citep{ramponi-2024-language}.

\paragraph{\textsc{Speaker demographic.}} 
The corpus consists of anonymized social media posts and therefore user demographics are unknown.

\paragraph{\textsc{Annotator demographic.}} 
The annotators are native speakers of Italian with background in linguistics and NLP. They identify themselves as a woman and a man, with age ranges 20--30 and 30--40. Both have naturally been exposed to public discourse around migration, climate change, and public health issues in the Italian context. They carried out data annotation as part of their work as employees at the host institution.

\paragraph{\textsc{Speech situation and text characteristics.}} 
The interaction is asynchronous and the speakers' intended audience is everyone. The genre of the written texts is social media with a focus on migration, climate change, and public health issues. The posts have been published within a 4-year time period (i.e., from \texttt{2019-01-01} to \texttt{2022-12-31}) and thus temporal biases in the dataset are minimized. The posts have been collected in February 2023.

\paragraph{\textsc{Preprocessing and data formatting.}} 
The posts have been anonymized by replacing user mentions, email addresses, phone numbers, and URLs with placeholders (i.e., \texttt{[USER]}, \texttt{[EMAIL]}, \texttt{[PHONE]} and \texttt{[URL]}, respectively). All emojis have been preserved in the text since they frequently signal \emph{language fallacies}, whereas newline characters (i.e., \texttt{\textbackslash n}, \texttt{\textbackslash r}) have been replaced with single spaces.

\section{Search Keywords} \label{app:search-keywords}

The search keywords have been selected from publicly-available glossaries, manuals, and reports produced by universities, agencies, and associations that deal with migration, climate change, and public health issues. Keyword selection was conducted ensuring to cover various subtopics, and each term/phrase was extended to cover all applicable grammatical genders and numbers. We report the original sources below and refer the reader to Table~\ref{tab:search-keywords} for the full set of keywords across topics.

\paragraph{Migration}

We use the \emph{United Nations High Commissioner for Refugees}' glossary in Italian~\citep{unhcr-2020-glossario} and the \emph{Guidelines for the Application of the Charter of Rome}~\citep{barretta-etal-2018-linee}. The resulting keywords represent the following aspects of migration: \emph{i)} phenomena, \emph{ii)} people, and \emph{iii)} status and hospitality (see Table~\ref{tab:search-keywords}).

\paragraph{Climate change} We rely on \emph{The Words of Climate Change}, a linguistic manual that includes the definition for over 200 concepts in Italian by 82 authors about climate change across 30 diverse subject areas~\citep{latini-etal-2020-lessico}. Those include climate change concepts that span environmental, climate, energy, chemical, physical, social, and economic subject areas, among others (see Table~\ref{tab:search-keywords}).

\paragraph{Public health} We mainly rely on the \emph{Health Promotion Glossary of Terms} by the~\citet{who-2021-health} by manually translating terms or finding corresponding translations in the Italian version of the glossary~\citep{barbera-etal-2012-glossario}. To broaden the scope of the search, we also draw keywords about specific public health areas from the \emph{Glossary on the Subject of Waiting Lists} by the Italian~\citet{ministry-2019-glossario}, from a glossary on health inequalities,\footnote{\url{https://health-inequalities.eu/resources/glossary/}} and from the ``Themes'' section of the Italian Ministry of Health website.\footnote{\url{https://www.salute.gov.it/portale/temi/p2_2.html}} The resulting keywords are in Table~\ref{tab:search-keywords}.

\begin{table*}
    \centering
    \resizebox{1.0\linewidth}{!}{%
    \begin{tabular}{p{2cm}p{16.5cm}}
        \toprule
        \textbf{Topic} & \textbf{Search keywords} \\
        \midrule
        \textsc{Migration} & 
        apolid[e,i]; 
        apolidia; 
        centr[o,i] di accoglienza; 
        centr[o,i] di identificazione ed espulsione; 
        centr[o,i] di permanenza per il rimpatrio; 
        centri di permanenza per i rimpatri; 
        centr[o,i] di permanenza temporanea; 
        centr[o,i] per il rimpatrio; 
        centri per i rimpatri; 
        corridio[io,i] umanitar[io,i]; 
        domand[a,e] d'asilo; 
        domand[a,e] di asilo; 
        emigrant[e,i]; 
        emigrat[o,i,a,e]; 
        emigrazion[e,i]; 
        espatr[io,i]; 
        fattor[e,i] di spinta; 
        immigrant[e,i]; 
        immigrat[o,i,a,e]; 
        immigrazion[e,i]; 
        ius sanguinis; 
        migrant[e,i]; 
        migrator[io,i,ia,ie]; 
        migrazion[e,i]; 
        minor[e,i] stranier[o,i] non accompagnat[o,i]; 
        minor[e,i] stranier[a,e] non accompagnat[a,e]; 
        non-refoulemen[t,ts]; 
        permess[o,i] di soggiorno; 
        procedur[a,e] d'asilo; 
        procedur[a,e] di asilo; 
        protezion[e,i] sussidiari[a,e]; 
        protezion[e,i] umanitari[a,e]; 
        push facto[r,rs]; 
        refoulemen[t,ts]; 
        reinsediament[o,i]; 
        respingiment[o,i]; 
        richiedent[e,i] asilo; 
        rifugiat[o,i,a,e]; 
        rimpatr[io,i]; 
        rimpatriat[o,i,a,e]; 
        sfollat[o,i,a,e]; 
        vittim[a,e] della tratta; 
        vittim[a,e] di tratta
        \\
        
        \midrule
        
        \textsc{Climate change} & 
        acidificazione dell'oceano; 
        acidificazione degli oceani; 
        aerosol atmosferic[o,i]; 
        allagament[o,i]; 
        alluvion[e,i]; 
        alluvional[e,i]; 
        ambientalismo di facciata; 
        anidride carbonica; 
        antropocene; 
        aridità; 
        bilanc[io,i] climatic[o,i]; 
        bilanc[io,i] energetic[o,i]; 
        bilanc[io,i] idrologic[o,i]; 
        biocombustibil[e,i]; 
        biodegradabil[e,i]; 
        biodegradabilità; 
        biodiversità; 
        biossido di carbonio; 
        cambiament[o,i] climatic[o,i]; 
        cambiament[o,i] del clima; 
        carbon cost; 
        carbon footprint; 
        carbon pricing; 
        carbon tax; 
        cost[o,i] del carbonio; 
        climate; 
        climate change; 
        climate cris[is,es]; 
        climatic[o,a,i,he]; 
        climatologia; 
        co2; 
        combustibil[e,i] fossil[e,i]; 
        confin[e,i] planetar[io,i]; 
        consum[o,i] di suolo; 
        crisi climatic[a,he]; 
        deforestazion[e,i]; 
        desalinizzazion[e,i]; 
        desertificazion[e,i]; 
        diossido di carbonio; 
        disboscament[o,i]; 
        dissalazion[e,i]; 
        ecological footprint; 
        ecologismo di facciata; 
        economi[a,e] circolar[e,i]; 
        effetto serra; 
        emission[e,i]; 
        energi[a,e] rinnovabil[e,i]; 
        esondazion[e,i]; 
        event[o,i] meteorologic[o,i] estrem[o,i]; 
        fenomen[o,i] meteorologic[o,i] estrem[o,i]; 
        finanza sostenibile; 
        fonte di energia rinnovabile; 
        fonti di energia rinnovabil[e,i]; 
        forzant[e,i] radiativ[o,i]; 
        gas serra; 
        gas silvestre; 
        glacialism[o,i]; 
        glaciazion[e,i]; 
        greenwashing; 
        impronta carbonica; 
        impronta di carbonio; 
        impronta ecologica; 
        innalzamento de[l,i] mar[e,i]; 
        innalzamento del livello de[l,i] mar[e,i]; 
        innalzamento dei livelli de[l,i] mar[e,i]; 
        inondazion[e,i]; 
        inquinamento atmo-sferico; 
        inquinamento dell'atmosfera; 
        isol[a,e] di calore; 
        isol[a,e] urban[a,e] di calore; 
        limit[e,i] planetar[io,i]; 
        meteorologia; 
        microclima; 
        mobilità sostenibile; 
        mutament[o,i] climatic[o,i]; 
        olocene; 
        ondat[a,e] di caldo; 
        ondat[a,e] di calore; 
        paleoclima; 
        particellato; 
        particolato; 
        pedoclima; 
        permafrost; 
        permagelo; 
        prezz[o,i] del carbonio; 
        proiezion[e,i] climatic[a,he]; 
        report di sostenibilità; 
        riscaldamento climatico; 
        riscaldamento globale; 
        risch[io,i] climatic[o,i]; 
        scenar[io,i] climatic[o,i]; 
        sciogliment[o,i] dei ghiacciai; 
        siccità; 
        sistem[a,i] climatic[o,i]; 
        sostenibilità ambientale; 
        surriscaldamento climatico; 
        surriscaldamento globale; 
        svilupp[o,i] sostenibil[e,i]; 
        tass[a,e] sul carbonio; 
        transizion[e,i] ecologic[a,he]; 
        transizion[e,i] energetic[a,he]; 
        uso d[el,i] suolo; 
        utilizzazion[e,i] del suolo; 
        utilizzo d[el,i] suolo; 
        variabilità climatic[a,he]
        \\
        
        \midrule
        
        \textsc{Public health} & 
        agend[a,e] di prenotazione; 
        alfabetizzazione alla salute; 
        alfabetizzazione sanitaria; 
        assistenz[a,e] domiciliar[e,i]; 
        assistenz[a,e] ospedalier[a,e]; 
        assistenz[a,e] sanitari[a,e]; 
        assistenza universale; 
        aziend[a,e] o-spedalier[a,e]; 
        aziend[a,e] sanitari[a,e]; 
        bisogn[o,i] sanitar[io,i]; 
        calendar[io,i] di prenotazione; 
        caric[o,hi] di malattia; 
        centro unificato di prenotazione; 
        città san[a,e]; 
        class[e,i] di priorità; 
        comportament[o,i] a rischio; 
        comportament[o,i] di salute; 
        copertur[a,e] sanitari[a,e]; 
        copertur[a,e] universal[e,i]; 
        cur[a,e] medic[a,he]; 
        cur[a,e] sanitari[a,e]; 
        degent[e,i]; 
        degenz[a,e]; 
        determinant[e,i] della salute; 
        determinant[e,i] di salute; 
        dimission[e,i] ospedalier[a,e]; 
        dispositiv[o,i] medic[o,i]; 
        disuguaglianz[a,e] di salute; 
        disuguaglianz[a,e] nella salute; 
        disuguaglianz[a,e] sanitari[a,e]; 
        educazione alla salute; 
        educazione sanitaria; 
        epidemi[a,e]; 
        epidemic[o,a,i,he]; 
        epidemiologia; 
        epidemiologic[o,a,i,he]; 
        equità di salute; 
        equità nella salute; 
        equità sanitari[a,e]; 
        esenzion[e,i] dal ticket; 
        esenzion[e,i] ticket; 
        fattor[e,i] di rischio; 
        indicator[e,i] di salute; 
        investiment[o,i] nella sanità; 
        investiment[o,i] per la salute; 
        investiment[o,i] per la sanità; 
        isol[a,e] san[a,e]; 
        istitut[o,i] di cura; 
        istituto di sanità pubblica; 
        istituto superiore di sanità; 
        list[a,e] di attesa; 
        malatti[a,e] infettiv[a,e]; 
        ministero della salute; 
        ministero della sanità; 
        misur[a,e] sanitari[a,e]; 
        ospedali; 
        ospedalier[o,i,a,e]; 
        ospeda-lizzazion[e,i]; 
        ospitalizzazion[e,i]; 
        pandemi[a,e]; 
        politic[a,he] sanitari[a,e]; 
        post[o,i] letto; 
        prestazion[e,i] ambulatorial[e,i]; 
        prestazion[e,i] sanitari[a,e]; 
        prestazion[e,i] specialistic[a,he] ambulatorial[e,i]; 
        prevenzione delle malattie; 
        prevenzione di malattie; 
        prevenzione primaria; 
        prevenzione sanitaria; 
        prevenzione secondaria; 
        prevenzione terziaria; 
        programmazion[e,i] sanitari[a,e]; 
        promozione della salute; 
        promozione di salute; 
        pronto soccorso; 
        ricover[o,i]; 
        salute globale; 
        salute per tutti; 
        salute pubblica; 
        sanità; 
        sanità pubblica; 
        sanitar[io,i,ia,ie]; 
        serviz[io,i] infermieristic[o,i]; 
        serviz[io,i] medic[o,i]; 
        serviz[io,i] sanitar[io,i]; 
        settor[e,i] sanitar[io,i]; 
        sicurezza dell[a,e] cur[a,e]; 
        struttur[a,e] di ricovero; 
        struttur[a,e] ospedalier[a,e]; 
        struttur[a,e] sanitari[a,e]; 
        terapi[a,e] intensiv[a,e]; 
        trattament[o,i] di salute; 
        trattament[o,i] medic[o,i]; 
        trattament[o,i] sanitar[io,i]; 
        uguaglianz[a,e] di salute; 
        uguaglianz[a,e] nella salute; 
        uguaglianz[a,e] sanitari[a,e]; 
        vaccin[o,i]; 
        vaccinazion[e,i]
        \\
        
	\bottomrule
	\end{tabular}
        }%
	\caption{\label{tab:search-keywords} Search keywords used for collecting posts about migration, climate change, and public health issues. We report grammatical gender and number variants (if any) using squared brackets. If more than one bracket is present for a term/phrase, the variants must be read by considering the elements with the same index within each bracket.}
\end{table*}


\section{Annotation Guidelines} \label{app:annotation-guidelines}

In this section, we present the guidelines we designed for the annotation of fallacies in \name. We first introduce the general guidelines (Section~\ref{app:annotation-guidelines-general}), i.e., those concerning neutrality and the identification of spans regardless of fallacy types. We then present fallacy-specific guidelines and extended definitions for each fallacy type (Section~\ref{app:annotation-guidelines-specific}).

\subsection{General Guidelines} \label{app:annotation-guidelines-general}

\paragraph{Neutrality in annotation} 
Fallacies must be identified on the basis of their argumentative invalidity, without considering the truth or falsity of the statement or whether the position advanced is ideologically agreeable. Any political or personal judgment must be set aside during annotation.

\paragraph{Extent of fallacy spans} 
We adopt a minimalist approach and annotate the smallest linguistic unit which expresses the fallacy. All overlapping fallacies must be annotated, regardless of negation. Four types of annotation spans can be identified:

\begin{enumerate}
    \item \textbf{Word level.} The fallacy is expressed through a single word, e.g.: ``il \textbf{becero} profitto'' (en: ``\emph{the \textbf{vulgar} profit}'') [\emph{Loaded language}];
    \item \textbf{Phrase level.} The fallacy is the union of more words, e.g.: ``\textbf{brigate rosse}'' (en: ``\emph{\textbf{red brigades}''}) [\emph{Name calling or labeling}]. Fallacies of the same type that occur close to each other must be annotated together, e.g.: ``\textbf{[che razza] di padre [abbietto]!}'' (en: ``\emph{\textbf{[what a] [despicable] father!}}'') [\emph{Loaded language}];
    \item \textbf{Clause level.} All the clause contributes to express the fallacy. To highlight the logical passage from the premise to the conclusion, we require the annotation of conjunctions, relative pronouns, and verbs, e.g.: ``questo \textbf{è il primo passo verso la dittatura}'' (en: ``\emph{this \textbf{is the first step towards dictatorship}}'') [\emph{Slippery slope}]; or ``sostieni che bisogna lavorare, \textbf{ma non eri proprio tu quello che faceva i festini a casa tua?}'' (en: ``\emph{You argue that people should work, \textbf{but weren't you the one throwing parties at your house?}}'') [\emph{Ad hominem}];
    \item \textbf{Sentence/post level.} The fallacy may be expressed through the whole content of the post and usually requires two arguments, e.g.: ``\textbf{la Dora è piena di acqua, non è vero che c'è il riscaldamento globale!}'' (en: ``\emph{\textbf{The Dora is full of water, so it's not true that there is global warming!}}'' [\emph{Cherry picking}]. If the premise or conclusion alone is sufficient, only the necessary part must be annotated (see individual fallacy guidelines in Appendix~\ref{app:annotation-guidelines-specific}). 
\end{enumerate}

\paragraph{Special characters}
Punctuation, emojis, emoticons, uppercase letters, and other graphic signs must be annotated when they carry semantic content that influences the argument or contribute to express the fallacy. Emojis and emoticons frequently convey fallacies such as \emph{Appeal to emotion} and \emph{Loaded language}. 
Uppercase letters and exclamation points are often used in the expression of the \emph{Loaded language} fallacy. When annotating punctuation, contiguous marks referring to different fallacies must be annotated separately. Hashtags should be annotated including the ``\texttt{\#}'' character. 
Some hashtags may simply serve as tags to ease the retrieval of posts about a topic and therefore do not express fallacies, whereas others are often used in \emph{Slogan} fallacies. Among other symbols, \texttt{[USER]} placeholders must be annotated if they are part of an \emph{Appeal to authority} fallacy. \texttt{[URL]} placeholders must instead be excluded from annotation.

\paragraph{Claims/arguments made by others}
Fallacies must be annotated even when part of the arguments are advanced by others. In this case, quotation marks should be excluded from annotation. Instead, reported testimonies do not require annotation since they typically convey personal opinions.

\paragraph{Pragmatic strategies}
Pragmatic strategies like irony can overlap with fallacies as they both involve violations of communicative norms~\citep{van2004systematic}. Fallacious reasoning should be annotated in contexts involving irony, following the European guidelines for annotation of persuasion techniques~\citep{piskorski-etal-2023-news}.

\subsection{Fallacy-specific Guidelines} \label{app:annotation-guidelines-specific}

We here provide the extended definitions for fallacies along with fallacy-specific guidelines (\faLightbulb). Examples for each fallacy are provided in Table~\ref{tab:examples}.

\paragraph{Ad hominem (\textsc{ah})}
A personal attack to an individual or a group which deviates from the main thesis. It comprises \emph{Abusive} \textsc{ah} (i.e., when the opponent's character is attacked), \emph{Circumstantial} \textsc{ah} (i.e., when the opponent is accused of being motivated by personal interest), and \emph{Tu quoque} (i.e., when there is a contradiction between what the opponent says or does and what they have said or done)~\citep{van2004systematic}.

\smallskip
\noindent \faLightbulb~~The attack can be addressed to a third person or group. The annotation must only include the attack itself, without the premises.

\paragraph{Appeal to authority (\textsc{aa})}
The author appeals to an authority or a group consensus to support their thesis, without further evidence. Following~\citet{goffredo2022fallacious}, we group two fallacies under this label, namely \emph{Appeal to authority} and \emph{Ad populum}/\emph{Bandwagon}. This fallacy also includes the cases in which the author appeals to their own authority or opinion.

\smallskip
\noindent \faLightbulb~~The annotation comprises the smallest segment in which the authority is mentioned, the declarative conjunction (or a colon), and additional information about the circumstances, such as the source.

\paragraph{Appeal to emotion (\textsc{ae})}
It involves the use of negative or positive personal emotions (e.g., shame, indignation, pity, or affection) to intentionally or unintentionally influence the audience. The label also includes the persuasion technique \emph{Appeal to fear/prejudice} as defined by~\citet{da-san-martino-etal-2019-fine}.

\smallskip
\noindent \faLightbulb~~This fallacy is a subtype of \emph{Red herring} as it deviates the attention from the main thesis, regardless of the language in use. In contrast, the subtype \emph{Loaded language} involves strong language use.

\paragraph{Causal oversimplification (\textsc{co})}
It involves a simplified and fallacious causal relation. It includes the subtypes \emph{False cause} and \emph{Post hoc} as defined by~\citet{musi2022developing}. 

\smallskip
\noindent \faLightbulb~~Both the premises and the conclusion must be annotated.

\paragraph{Cherry picking (\textsc{cp})}
This fallacy consists of choosing evidence to support a thesis while ignoring any other contrary evidence~\citep{musi2022developing}.

\smallskip
\noindent \faLightbulb~~In contrast to \emph{Hasty generalization}, where the author builds a conclusion based on the evidence, here the author selects evidence that supports a pre-existing conclusion. Both the claim that confirms the thesis and the thesis itself must be annotated.

\paragraph{Circular reasoning (\textsc{cr})}
An error of circularity: the end of an argument comes back to the beginning without having proven itself~\citep{jin-etal-2022-logical}. 

\smallskip
\noindent \faLightbulb~~The entire argumentation includes the premises and the conclusion, which usually extend over multiple sentences, and that must be annotated.

\paragraph{Doubt (\textsc{do})}
It is used to intentionally question the credibility of someone or something~\citep{da-san-martino-etal-2019-fine}.

\smallskip
\noindent \faLightbulb~~It usually involves linguistic devices such as question marks, adverbs of doubt (e.g.: ``\textbf{forse}'', en: ``\emph{maybe}''; ``\textbf{magari}'', en: ``\emph{perhaps}''), adversative conjunctions (e.g.: ``\textbf{ma}'', en: ``\emph{but}''; ``\textbf{però}'', en: ``\emph{however}''), conditional conjunctions (e.g.: ``\textbf{se}'', en: ``\emph{if}''), and rhetorical questions (e.g.: ``\textbf{siamo sicuri che...?}'', en: ``\emph{are we sure that...?}''; ``\textbf{perché non...?}'', en: ``\emph{why not...?}'').

\paragraph{Evading the burden of proof (\textsc{ep})}
A thesis is advanced without any support as if it was self-evident, meaning that one or more arguments are missing in the argument structure~\citep{musi2022developing}. 

\smallskip
\noindent \faLightbulb~~The fallacy should not be annotated in cases when no evidence could be theoretically provided. URLs can sometimes contain more information about the statement, neutralizing the fallacy.

\paragraph{False analogy (\textsc{fa})}
This fallacy occurs when two different things or situations are placed on the same level because they are supposed to share similar aspects. The label includes \emph{False analogy} as defined by~\citet{musi2022developing} and \emph{False equivalence} as defined by~\citet{phillips2012purposeful}.

\smallskip
\noindent \faLightbulb~~It can also include lists of entities that are implicitly equated, e.g.: ``\textbf{migranti e spazzatura}'' (en: ``\emph{migrants and trash}''). The annotated span must include both/all the concepts being discussed.

\paragraph{False dilemma (\textsc{fd})}
It presents only two options or sides when there are many~\citep{jin-etal-2022-logical}.

\smallskip
\noindent \faLightbulb~~The annotated span must include both the things being presented.

\paragraph{Flag waving (\textsc{fw})}
This fallacy occurs when the author intentionally plays on a sense of belonging to a country, a group, or an ideology to support an argument, as if waving a flag. 

\smallskip
\noindent \faLightbulb~~The fallacy can be frequently found in hashtags. Examples include names of political parties (\textit{e.g.: \#FratellidItalia}), or manifestations and organizations (\textit{e.g.: \#FridaysForFuture}).

\paragraph{Hasty generalization (\textsc{hg})}
It occurs when a generalization is drawn from a sample which is too small, not representative, or not applicable to the whole situation if all the variables are taken into account~\citep{musi2022developing}. It is an example of \emph{Jumping to conclusions}~\citep{jin-etal-2022-logical}. 

\smallskip
\noindent \faLightbulb~~We require to annotate the generalization itself, not the premises. In some cases, the generalization just consists in an overly broad statement, where the generalized sample is not expressed.

\paragraph{Loaded language (\textsc{ll})}
It involves using words or phrases with strong emotional implications (either positive or negative) to influence the audience~\citep{da-san-martino-etal-2019-fine}. 

\smallskip
\noindent \faLightbulb~~Cases that are not functional to the argumentation, regardless of intentionality, are not annotated. This fallacy is often found at word level. Frequent devices are swear words (e.g.: ``\textbf{cazzo}'', en: ``\emph{shit}''), slang (e.g.: ``\textbf{sbirri}'', en: ``\emph{cops}''), evaluative terms (e.g.: ``\textbf{famigerati}'', en: ``\emph{infamous}''), colloquial expressions (e.g.: ``\textbf{non frega nulla}'', en: ``\emph{they don't care}''), discourse markers (e.g.: ``\textbf{dai}'', en: ``\emph{come on}''; ``\textbf{ovviamente}'', en: ``\emph{obviously}''), rhetorical strategies (e.g.: ``\textbf{ma per piacere...}'', en: ``\emph{oh, please...}''), and repetitions, as well as graphic strategies like emojis, emoticons, hashtags, punctuation (e.g., exclamation marks), or uppercase letters. Instances which do not have inherent connotations but acquire a connotation in the pragmatic context should be evaluated individually by the annotator.

\paragraph{Name calling or labeling (\textsc{nc})}
This fallacy involves labeling something or someone positively or negatively to influence the audience, for example associating it with an ideology. We group under this label the two propaganda techniques \textit{Name calling or labeling} and \textit{Reductio ad Hitlerum} as defined by~\citet{da-san-martino-etal-2019-fine}.

\smallskip
\noindent \faLightbulb~~The annotation includes the target, the label, and the article. In contrast with \emph{Loaded language}, here the linguistic device specifically brings the target back to an ideological or minority group in which they do not identify.

\paragraph{Red herring (\textsc{rh})}
The argument supporting the claim diverges the attention to issues which are irrelevant for the claim at hand~\citep{musi2022developing}. It includes the subtypes \textit{Appeal to worse problems}, \textit{Appeal to tradition}, and \textit{Appeal to nature} as defined in~\citet{sahai-etal-2021-breaking}.

\smallskip
\noindent \faLightbulb~~Only the passage that deviates from the thesis must be annotated.

\paragraph{Slippery slope (\textsc{ss})}
It implies that an exaggerated consequence could result from a particular action \citep{goffredo-etal-2023-argument}. 

\smallskip
\noindent \faLightbulb~~The annotation must only include the fallacy itself, without the premises. Note that, if the exaggeration is a plausible fact supported by evidence, it should not be annotated. 

\paragraph{Slogan (\textsc{sl})}
It consists of a brief and striking phrase that is used to provoke excitement of the audience~\citep{goffredo-etal-2023-argument}.

\smallskip
\noindent \faLightbulb~~This fallacy can be frequently found in hashtags. Each hashtag is annotated separately, except when a sequence of related hashtags is found. In contrast to \emph{Flag waving}, it does not specifically plays on a sense of belonging to a group or ideology. Examples include \textit{\#ClimateChangeIsReal}, \textit{\#RiseForClimate}, \textit{\#ActOnClimate}. 

\paragraph{Strawman (\textsc{st})}
It consists of distorting someone else's argument and then tearing it down. The arguer misinterprets an opponent’s argument for the purpose of more easily attacking it, demolishes it, and then concludes that the opponent’s real argument has been demolished \citep{musi2022developing}.

\smallskip
\noindent \faLightbulb~~We require the annotation of the reported argument, which frequently includes \textit{Vagueness}, and its attack.

\paragraph{Thought-terminating cliché (\textsc{tc})}
It consists of a short and generic phrase that discourages critical thought and meaningful discussion~\citep{da-san-martino-etal-2019-fine}.

\smallskip
\noindent \faLightbulb~~It is usually found at the end of a sentence and involves a final punctuation mark that should be annotated. Phrases created ad hoc by the post author must also be annotated, including corner cases such as ``\textbf{ma azzardo}'' (en: ``\emph{just guessing}''). 

\paragraph{Vagueness (\textsc{va})}
It is found when ambiguous words are shifted in meaning in the process of arguing or are left vague, being potentially subject to skewed interpretations~\citep{musi2022developing}. The label also includes the \textit{Equivocation} fallacy as defined by~\citet{jin-etal-2022-logical}.

\smallskip
\noindent \faLightbulb~~When the intentionality of the arguer is not clear, the annotator should evaluate if the content can be misinterpreted and lead to ambiguity. This often occurs with indefinite expressions such as ``\textbf{molti}'' (en: ``\emph{many}''), ``\textbf{quasi}'' (en: ``\emph{almost}''), ``\textbf{circa}'' (en: ``\emph{about}''), and general extenders such as ``\textbf{ecc.}'' (en: ``\emph{etc.}''). Vague expressions that do not play a role in influencing the audience and do not lead to potential ambiguity are not considered instances of \emph{Vagueness}, and therefore should not be annotated (e.g.: ``\textbf{dopo mesi...}'', en: ``\emph{after months...}'').

\begin{table*}[th!]
  \centering
  \resizebox{1.0\linewidth}{!}{%

  \begin{tabular}{ll}
    \toprule
     \textbf{Fallacy type} & \textbf{Example} \\
    \midrule
    Ad hominem & \makecell[cl]{\textbf{Gli accoglienti e umanitari francesi non vogliono più i \#migranti della}\\\textbf{\#SeaWatch3. La loro parola vale meno di una scoreggia}.\\\emph{\textbf{The welcoming and humanitarian French no longer want the \#migrants on the}}\\\emph{\textbf{\#SeaWatch3. Their word is worth less than a fart}.}}\\
    \midrule
    Appeal to authority & \makecell[cl]{\textbf{Lo dice anche la \#Bundesbank}, i \#migranti servono ad abbassare i salari di tutti.\\\emph{\textbf{The \#Bundesbank also says it}, \#migrants serve to lower everyone's wages.}} \\
    \midrule
    Appeal to emotion & \makecell[cl]{La manifestazione a \#Voghera è \textbf{arrabbiata, dignitosa e ordinata}. Sta girando per\\tutta la città con molti immigrati e l'appoggio della sinistra. \textbf{Vuole giustizia.}\\\emph{The demonstration in \#Voghera is \textbf{angry, dignified, and orderly}. It is moving}\\\emph{through the city with many immigrants and the support of the left. \textbf{It wants justice.}}} \\
    \midrule
    Causal oversimplification & \makecell[cl]{\textbf{Ho fatto lavare l'auto. Dopo mesi di siccità, sono sicuro che pioverà presto}.\\\emph{\textbf{I had the car washed. After months of drought, I'm sure it will rain soon}.}} \\
    \midrule
    Cherry picking & \makecell[cl]{\textbf{Questo video rinfrescherà la memoria a chi l'ha corta e mostrerà ai giovani che}\\\textbf{Roma ha sempre avuto voragini e alberi caduti dopo forti piogge}. \\\emph{\textbf{This video will refresh the memory of those with short memories and reveal to the}} \\\emph{\textbf{young that Rome has always had sinkholes and fallen trees after heavy rains}.}} \\
    \midrule
    Circular reasoning & \makecell[cl]{\textbf{Le leggi sull'immigrazione sono già così fasciste che quando i fascisti}\\\textbf{promettono di introdurre nuove misure, stanno solo descrivendo le norme}\\\textbf{fasciste che già esistono e sono applicate da decenni}.\\\emph{\textbf{Immigration laws are already so fascist that when fascists promise to introduce}}\\\emph{\textbf{new measures, they're actually just describing the fascist rules that already}}\\\emph{\textbf{exist and have been applied for decades}.}} \\
    \midrule
    Doubt & \makecell[cl]{Nella strage di Lampedusa morirono 368 migranti. L'Italia e l'Europa dissero ``mai \\più''. \textbf{Forse ci credevano davvero}.\\\emph{In the Lampedusa massacre, 368 migrants died. Italy and Europe said ``never again''.}\\\emph{\textbf{Maybe they really believed it}.}}\\
    \midrule
    Evading the burden of proof & \makecell[cl]{Trovo geniali \textbf{tedeschi e svedesi che censurano i crimini degli immigrati per}\\\textbf{restare in testa alla classifica dei popoli meno xenofobi}.\\\emph{I find it ingenious how \textbf{Germans and Swedes censor immigrants' crimes to stay}}\\\emph{\textbf{ahead in the ranking of the least xenophobic nations}.}} \\
    \midrule
    False analogy & \makecell[cl]{Non vi fa sorgere qualche dubbio \textbf{uno stato che blocca le \#navidacrociera per} \\\textbf{prevenire il \#covid ma fa sbarcare centinaia di immigrati dalle navi delle \#ONG}\\\textbf{e dai \#barchini}? \\\emph{Doesn't it raise some doubts about \textbf{a state that blocks \#cruiseships to prevent \#covid,}}\\\emph{\textbf{but allows hundreds of immigrants to disembark from \#ONG ships and \#smallboats}?}}\\
    \midrule
    False dilemma &  \makecell[cl]{\textbf{Queste elezioni saranno uno spartiacque tra chi mette al centro la sostenibilità}\\ \textbf{ambientale e chi difende il combustibile fossile}.\\\emph{\textbf{These elections will be a turning point between those who prioritize environmental}}\\\emph{\textbf{sustainability and those who defend fossil fuels}.}}\\
    \midrule
    Flag waving &  \makecell[cl]{\textbf{Con la forza e l'unità di questo popolo orgoglioso} chiediamo a tutta la \#Toscana di\\schierarsi subito. \textbf{Vinciamo, insieme, per la Toscana più forte e unita!}\\\emph{\textbf{With the strength and unity of these proud people}, we ask all of \#Tuscany to stand} \\\emph{with us immediately. \textbf{Let's win, together, for a stronger and more united Tuscany!}}}\\
    \midrule
    Hasty generalization & \makecell[cl]{Se Draghi ha mentito agli Italiani sul vaccino, \textbf{allora sta mentendo anche su guerra,}\\\textbf{clima, transizione ecologica, pnrr, energia, benzine e bollette}! \\\emph{If Draghi lied to Italians about the vaccine, \textbf{then he is also lying about the war, climate,}}\\\emph{\textbf{ecological transition, pnrr, energy, petrol, and bills}!}}\\
    \midrule
    Loaded language & \makecell[cl]{Epidemia colposa. \textbf{Bomba} in arrivo dalla Procura, \textbf{valanga} di indagati: \textbf{tremano}\\Speranza e Lorenzin. \\\emph{Culpable epidemic. \textbf{Bombshell} coming from the Public Prosecutor's Office, \textbf{avalanche}}\\\emph{of suspects: Speranza and Lorenzin \textbf{tremble}.}} \\
    \bottomrule \\[-5px]
    \multicolumn{2}{c}{\textcolor{gray}{\textbf{\emph{(Continued on the next page)}}}} \\
  \end{tabular}
  }%
\end{table*}

\begin{table*}[!th]
  \centering
  \resizebox{1.0\linewidth}{!}{%

  \begin{tabular}{ll}
  \multicolumn{2}{c}{\textcolor{gray}{\textbf{\emph{(Continued from the previous page)}}}} \\[5px]
    \toprule
     \textbf{Fallacy type} & \textbf{Example} \\
    \midrule
    Name calling or labeling & \makecell[cl]{\textbf{Un idiota fasciorazzista} condivide una foto di un giovane immigrato inventando fosse\\senza biglietto; foto condivisa 80mila volte con frasi \textbf{razziste}; smentita di Trenitalia:\\``Aveva solo sbagliato posto''. Siete solo dei \textbf{razzisti di merda}. \\\emph{\textbf{A fasci-racist idiot} shares a photo of a young immigrant claiming he was ticketless;} \\\emph{photo shared 80k times with \textbf{racist} comments; Trenitalia's denial: ``He just sat in the}\\\emph{wrong seat''. You're just \textbf{shitty racists}.}} \\
    \midrule
    Red herring & \makecell[cl]{\textbf{Siccome pandemia, crisi energetica, crisi climatica, inflazione e tensioni tra Russia}\\\textbf{e Ucraina non sono sufficienti}, oggi il Corriere ci regala questo articolo \coldface~``Fidati dei\\professionisti dell’informazione''.\\\emph{\textbf{Since pandemic, energy crisis, climate crisis, inflation, and tensions between Russia}}\\\emph{\textbf{and Ukraine aren't enough}, today the Corriere gives us this article} \coldface~``Trust the \\\emph{professionals of information''}.} \\
    \midrule
    Slippery slope & \makecell[cl]{[USER] dà la colpa delle violenze di Peschiera agli italiani. \textbf{Un altro vergognoso}\\\textbf{gradino verso la voragine della stampa italiana}. \\\emph{[USER] blames Italians for the violence in Peschiera. \textbf{Another shameful step towards}}\\\emph{\textbf{the abyss of Italian press}.}} \\
    \midrule
    Slogan & \makecell[cl]{La difesa dei confini e la lotta all'immigrazione incontrollata di massa rimarranno una\\priorità. \textbf{Difendiamo i confini!   \#BloccoNavaleSubito}\\\emph{The defense of borders and the fight against uncontrolled mass immigration will}\\\emph{remain a priority. \textbf{Let's defend the borders! \#NavalBlockadeNow}}} \\
    \midrule
    Strawman & \makecell[cl]{\textbf{L'incipit del TG1: ``Complice il bel tempo, continuano gli sbarchi a Lampedusa''.}\\\textbf{Vogliamo il mare a forza 8 così ci pensa lui a smaltire i migranti?!}\\\emph{\textbf{The incipit of TG1: ``Thanks to the good weather, landings in Lampedusa continue''.}}\\\emph{\textbf{Do we want a force 8 sea so it can get rid of migrants?!}}} \\
    \midrule
    Thought-terminating cliché &  \makecell[cl]{Nella UE tornano i muri antimigranti. \textbf{Era ora.}\\\emph{In EU, anti-migrant walls come back. \textbf{It was about time.}}} \\
    \midrule
    Vagueness & \makecell[cl]{Ogni centesimo speso in armi è tolto a sanità, ricerca, istruzione e transizione\\energetica: perciò non sono pro-Putin, che in questo \textbf{ha fatto una scelta ben precisa}.\\\emph{Every cent spent on weapons is taken away from healthcare, research, education, and}\\\emph{energy transition: therefore, I am not pro-Putin, who\textbf{ has made a very clear choice}}\\\emph{in this regard.}} \\
    \bottomrule
  \end{tabular}
  }%
  \caption{\label{tab:examples} Examples of annotated spans for each fallacy type. Annotated spans are indicated in bold and English translations are in italic. Examples have been slightly edited to preserve users' anonymity.}
\end{table*}


\section{Additional Dataset Details} \label{sec:additional-dataset-details}

\begin{figure*}[!ht]
\centering
\begin{subfigure}{.497\textwidth}
  \centering
  \includegraphics[width=.975\linewidth]{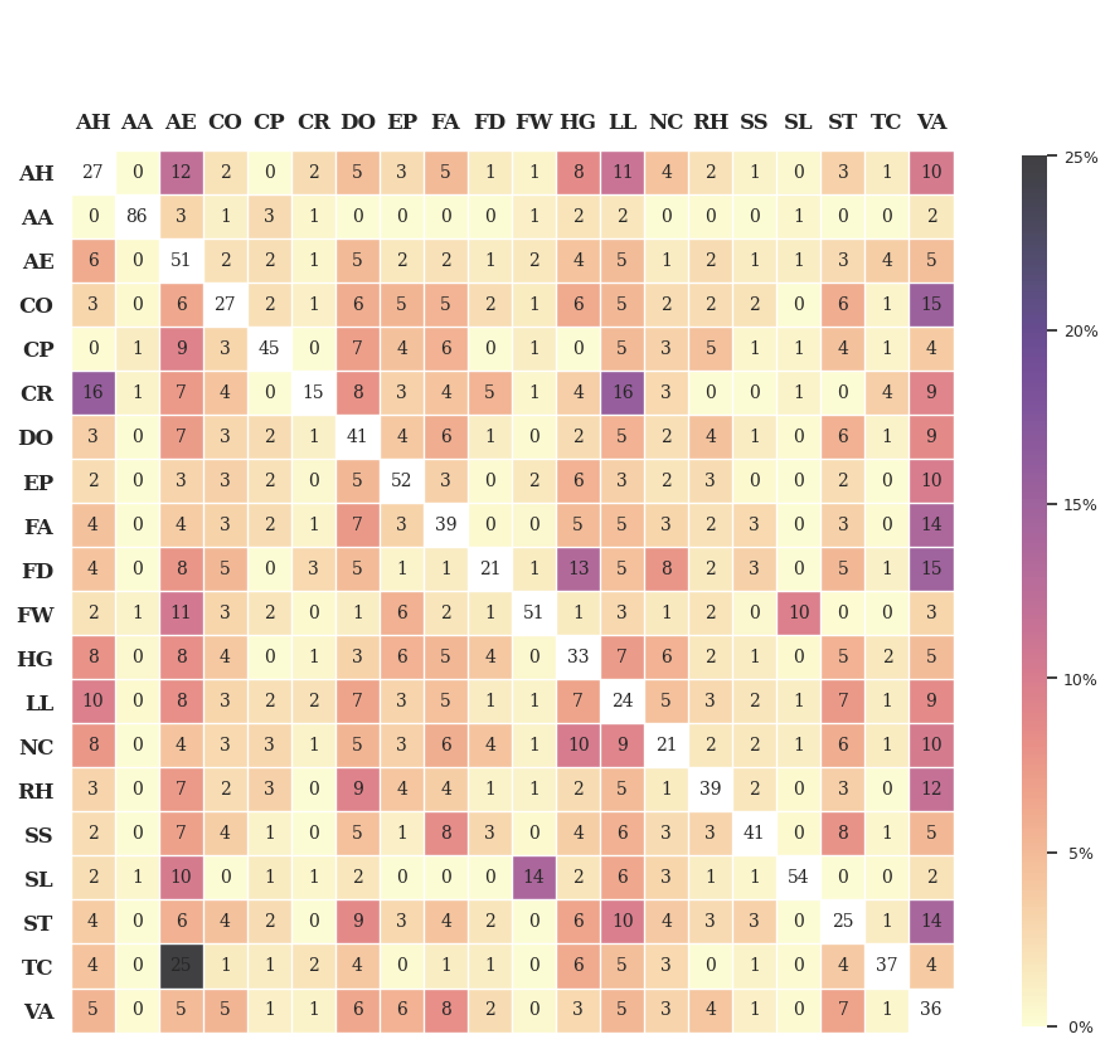}
  \caption{Fallacy overlaps in $\mathcal{A}_1$ annotations.}
  \label{fig:overlaps-a1}
\end{subfigure} \hfill
\begin{subfigure}{.497\textwidth}
  \centering
  \includegraphics[width=.975\linewidth]{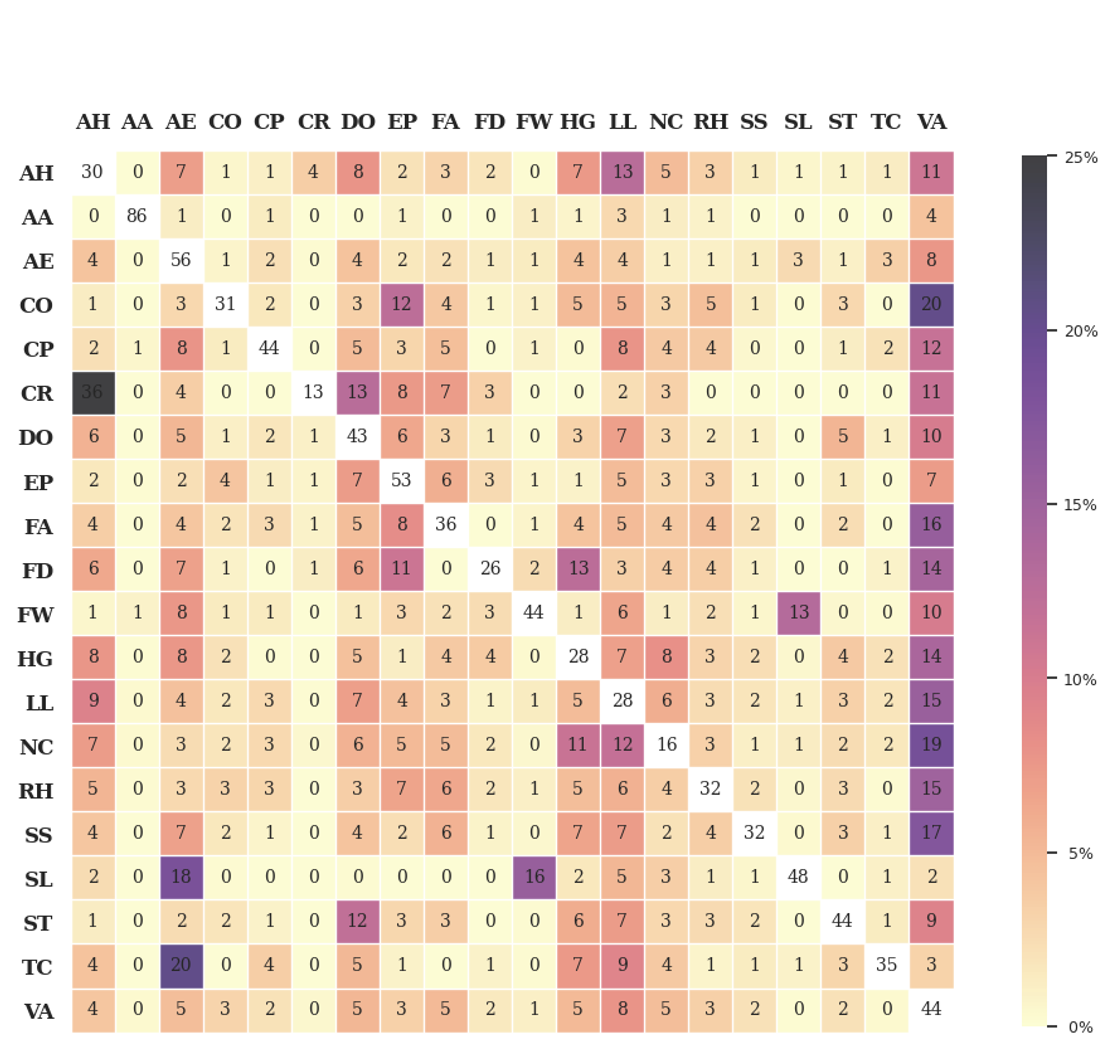}
  \caption{Fallacy overlaps in $\mathcal{A}_2$ annotations.}
  \label{fig:overlaps-a2}
\end{subfigure}
\caption{Overlap of fallacy annotations in $\mathcal{A}_1$ and $\mathcal{A}_2$ in terms of token percentages. Each row indicates the percentage of tokens for a given fallacy type that overlaps with any other fallacy type (\emph{columns}). White cells (\emph{diagonal}) indicate the percentage of tokens for each fallacy type that does not overlap with any other fallacy type. The overlap of fallacy tokens considering all the annotations ($\mathcal{A}_1 + \mathcal{A}_2$) is presented in Figure~\ref{fig:overlaps-full}.}
\label{fig:overlaps-individual}
\end{figure*}

In this section, we present additional details about the \name~dataset with regards to individual annotators' labels. Individual summary statistics are presented in Table~\ref{tab:stats-full}, whereas fallacy overlaps for each annotator are in Figure~\ref{fig:overlaps-individual}. 
We also use Variationist (v0.1.4)~\citep{ramponi-etal-2024-variationist} and calculate the top-$k$ ($k=10$) most informative tokens for fallacy types that have an average span length of $\leq10$ tokens (Table~\ref{tab:stats-full}),\footnote{For the analysis, we set both \texttt{lowercase} and \texttt{stopwords} to \texttt{True} and set the \texttt{language} to Italian (\texttt{it}). We employ normalized, positive, and weighted PMI as a metric.} i.e., those that are mainly related to language use and are therefore characterized by specific lexical choices. Results are in Table~\ref{tab:variationist}.

\begin{table*}[!ht]
  \centering
  \begin{tabular}{lrr|rr|rr}
    \toprule
     & \multicolumn{2}{c|}{\textbf{$\mathcal{A}_1 + \mathcal{A}_2$}} & \multicolumn{2}{c|}{\textbf{$\mathcal{A}_1$}} & \multicolumn{2}{c}{\textbf{$\mathcal{A}_2$}} \\
     \textbf{Fallacy type} & \textbf{Spans} & \textbf{Length} & \textbf{Spans} & \textbf{Length} & \textbf{Spans} & \textbf{Length} \\
    \midrule
    Ad hominem & 319 & 16.0$_{\pm 13.3}$ & 166 & 15.5$_{\pm 13.3}$ & 153 & 16.5$_{\pm 13.3}$ \\
    Appeal to authority & 213 & 6.4$_{\; \: \pm 4.3}$ & 120 & 6.4$_{\; \: \pm 4.5}$ & 93 & 6.4$_{\; \: \pm 3.9}$ \\
    Appeal to emotion & 2,049 & 5.1$_{\; \: \pm 4.9}$ & 1,022 & 5.7$_{\; \: \pm 5.6}$ & 1,047 & 4.5$_{\; \: \pm 4.0}$ \\
    Causal oversimplification & 142 & 19.0$_{\pm 10.7}$ & 87 & 19.4$_{\pm 11.2}$ & 55 & 18.3$_{\pm 10.0}$ \\
    Cherry picking & 94 & 28.8$_{\pm 12.3}$ & 52 & 28.2$_{\pm 12.5}$ & 42 & 29.5$_{\pm 12.0}$ \\
    Circular reasoning & 20 & 26.8$_{\pm 11.0}$ & 12 & 27.4$_{\pm 10.5}$ & 8 & 25.8$_{\pm 11.8}$ \\
    Doubt & 482 & 16.1$_{\pm 11.4}$ & 236 & 16.8$_{\pm 12.2}$ & 246 & 15.4$_{\pm 10.4}$ \\
    Evading the burden of proof & 406 & 16.2$_{\; \: \pm 9.9}$ & 195 & 16.3$_{\pm 10.7}$ & 211 & 16.2$_{\; \: \pm 9.2}$ \\
    False analogy & 239 & 22.1$_{\pm 13.4}$ & 147 & 20.8$_{\pm 13.0}$ & 92 & 24.1$_{\pm 13.8}$ \\
    False dilemma & 90 & 15.9$_{\pm 11.0}$ & 44 & 16.6$_{\pm 12.1}$ & 46 & 15.2$_{\; \: \pm 9.8}$ \\
    Flag waving & 393 & 4.3$_{\; \: \pm 4.9}$ & 197 & 4.6$_{\; \: \pm 5.5}$ & 196 & 3.9$_{\; \: \pm 4.2}$ \\
    Hasty generalization & 464 & 11.2$_{\; \: \pm 8.0}$ & 241 & 11.8$_{\; \: \pm 8.8}$ & 223 & 10.6$_{\; \: \pm 6.9}$ \\
    Loaded language & 2,484 & 2.5$_{\; \: \pm 2.7}$ & 1,064 & 2.8$_{\; \: \pm 3.6}$ & 1,420 & 2.4$_{\; \: \pm 1.6}$ \\
    Name calling or labeling & 1,124 & 2.6$_{\; \: \pm 1.7}$ & 561 & 2.5$_{\; \: \pm 1.6}$ & 563 & 2.7$_{\; \: \pm 1.7}$ \\
    Red herring & 257 & 13.0$_{\; \: \pm 8.5}$ & 129 & 14.2$_{\; \: \pm 9.3}$ & 128 & 11.8$_{\; \: \pm 7.4}$ \\
    Slippery slope & 172 & 10.8$_{\; \: \pm 6.8}$ & 88 & 11.4$_{\; \: \pm 7.5}$ & 84 & 10.2$_{\; \: \pm 6.0}$ \\
    Slogan & 384 & 3.5$_{\; \: \pm 3.1}$ & 190 & 3.4$_{\; \: \pm 3.0}$ & 194 & 3.5$_{\; \: \pm 3.2}$ \\
    Strawman & 109 & 36.3$_{\pm 15.4}$ & 67 & 34.9$_{\pm 16.1}$ & 42 & 38.5$_{\pm 14.0}$ \\
    Thought-terminating cliché & 285 & 5.2$_{\; \: \pm3.0}$ & 145 & 5.3$_{\; \: \pm 3.2}$ & 140 & 5.2$_{\; \: \pm 2.8}$ \\
    Vagueness & 1,338 & 9.1$_{\; \: \pm 8.6}$ & 536 & 9.5$_{\; \: \pm 9.3}$ & 802 & 8.9$_{\; \: \pm 8.2}$ \\
    \midrule
    \textbf{All} & 11,064 & 7.6$_{\; \: \pm 9.3}$ & 5,279 & 8.2$_{\; \: \pm 9.9}$ & 5,785 & 7.1$_{\; \: \pm 8.6}$ \\
    \bottomrule
  \end{tabular}
  \caption{\label{tab:stats-full} Per-annotator and overall annotation statistics across all fallacy types. We report the number of \emph{spans} and their average \emph{length} (with standard deviation) at the token level. $\mathcal{A}_1$ and $\mathcal{A}_2$ indicate annotation statistics by the individual annotators, whereas $\mathcal{A}_1 + \mathcal{A}_2$ summarizes the overall ones.}
\end{table*}

\begin{table*}[!h]
    \centering
    \resizebox{1.0\linewidth}{!}{%
    \begin{tabular}{ll}
        \toprule
        \textbf{Fallacy type} & \textbf{Top-$k$ tokens} \\
        \midrule
        Appeal to authority & \texttt{user}, articolo, studio, video, sa, papafrancesco, via, scientifico, intervista, scritto \\[2pt]
        Appeal to emotion & \clappinghands, \facewithtears, \redheart, schifo, \foldedhands, \facewithsteam, vergogna, \flexedbiceps, \greenheart, umanità \\[2pt]
        Flag waving & \makecell[cl]{italiani, \texttt{user}, fridaysforfuture, \flagitaly, m5s, \starr, fratelliditalia, italiasulserio, \\ fridayforfuture, piazza} \\[2pt]
        Loaded language & lotta, cazzo, leggi, merda, \redsign, caro, invasione, coglioni, combattere, disastro \\[2pt]
        Name calling or labeling & \makecell[cl]{immigrati, immigrazione, migranti, clandestini, sinistra, vax, negazionisti, \\ dittatura, immigrato, sanitaria} \\[2pt]
        Slogan & \makecell[cl]{primadeldiluvio, italiasulserio, facciamorete, vaccinare, sostenibile, \\ cambiaeninonilclima, climateactionnow, resistenza, toscana, iovotocalenda} \\[2pt]
        Thought-terminating cliché & mah, so, sapevatelo, eh, capita, semplice, vince, ciao, stop, ragione \\[2pt]
        Vagueness & \makecell[cl]{migranti, immigrati, immigrazione, lavoro, sanità, governo, politici, politica, \\ italiani, pandemia} \\
	\bottomrule
	\end{tabular}
        }%
	\caption{\label{tab:variationist} Top-$k$ ($k=10$) most informative tokens for fallacy types with an average length of $\leq10$ tokens, i.e., those that are mainly related to language use, considering all the annotations in the dataset ($\mathcal{A}_1 + \mathcal{A}_2$).}
\end{table*}


\section{Additional Experimental Details} \label{app:additional-experimental-details}

In this section, we describe additional details on our experimental settings. All our experiments were run on a single GPU (Tesla V100-SXM2-32GB).

\paragraph{Hyper-parameters} For supervised models, we employ the default MaChAmp~\citep{van-der-goot-etal-2021-massive} hyperparameter values (see Table~\ref{tab:hp}) and fine-tune the models for \textsc{span} and \textsc{post} tasks for \texttt{20} and \texttt{10} epochs, respectively. 
The AlBERTo and UmBERTo versions we used are \texttt{bert\_uncased\_L-12\_H-768\_A-12\_italian\_} \texttt{alb3rt0} and \texttt{umberto-commoncrawl-cased-v1}.
For unsupervised models, we use default settings from the Hugging Face library.

\begin{table}[!h]
    \centering
    \begin{tabular}{lr}
        \toprule
        \textbf{Hyperparameter} & \textbf{Value} \\
        \midrule
        Optimizer & AdamW \\ 
        $\beta_1$, $\beta_2$ & 0.9, 0.99 \\ %
        Dropout & 0.2 \\ %
        Epochs & 10 / 20 \\ %
        Batch size & 32 \\ %
        Learning rate & 1e-4 \\ %
        LR scheduler & Slanted triangular \\ %
        Decay factor & 0.38 \\ %
        Cut fraction & 0.3 \\ %
        \midrule
        Task loss weight ($\lambda$) & 1 \\
        Multi-label threshold ($\tau$) & 0.7 \\
		\bottomrule
	\end{tabular}
	\caption{\label{tab:hp} Hyperparameter values employed for the supervised models in all our experiments.}
\end{table}

\paragraph{Prompts and technical details}
The prompt template used across all experiments along with task-specific prompt variables are presented in Table~\ref{tab:prompt-vars}. 
For \textsc{span} tasks, the output format was initially requested in the CoNLL format with the \textsc{bio}-tagging scheme for unique identification of the fallacy segments; however, the models struggled to provide consistent outputs. Therefore, we provide an incremental identifier for each token in the input text (i.e., token id) and request the output following the format \texttt{[first number-last number = Fallacy Label]}. We chose square brackets because we observed that models could easily replicate them. Moreover, they facilitate the retrieval of the portion of the output in which fallacy predictions are actually provided, while disregarding other fallacy mentions across the output. We use regular expressions for extracting predictions from the output and also normalize the predicted fallacy names that clearly refer to the same fallacy type (e.g., we consider both the British spelling \emph{Name calling or labelling} and the shortened \emph{Name calling} label as instances of the \emph{Name calling or labeling} fallacy).

\paragraph{Results on individual test sets} In Table~\ref{tab:results-individual}, we present the full results obtained by each model across all task setups on individual test sets (i.e., $\mathcal{A}_1$ and $\mathcal{A}_2$), as well as those averaged over all $|\mathcal{A}|$ test set versions (i.e., $\mathcal{A}_1 + \mathcal{A}_2$).	

\begin{NewBox}*
\emph{Given an Italian text, your task is to \texttt{\$TASK\_DESC}. The 20 fallacy labels are as follows:}
\\ \\
\texttt{\$FALLACIES} + \texttt{\$MACROCAT}
\\ \\
\texttt{\$INSTR\_1} + \texttt{\$INSTR\_2} \\
\emph{There’s no limit in the number of fallacies you can find in the text, so it’s really important that you identify all possible fallacies.} \\
\texttt{\$INSTR\_3}
\\ \\
\emph{Here is the Italian text you must analyze} + \texttt{\$INSTR\_4}:
\\
\texttt{\$INPUT}
\\ \\
\texttt{\$INSTR\_5}
\\
\emph{Now return the output for the provided text.}
\end{NewBox}

\begin{table*}
    \centering
    \resizebox{1.0\linewidth}{!}{%
    \begin{tabular}{l|cc|cc|l}
        \toprule
        \textbf{Variable} & \textbf{\textsc{span}} & \textbf{\textsc{post}} & \textbf{\textsc{f}} & \textbf{\textsc{c}} & \textbf{Text snippet} \\
        \midrule
        \texttt{\$TASK\_DESC} & \faCheckSquare & & \faCheckSquare & \faCheckSquare & \emph{detect and classify the segments of text that contain fallacies} \\
        \texttt{\$TASK\_DESC} & & \faCheckSquare & \faCheckSquare & \faCheckSquare & \emph{detect the fallacies that are expressed in it} \\
        \midrule
        \texttt{\$FALLACIES} & \faCheckSquare & \faCheckSquare & \faCheckSquare & \faCheckSquare & ``\texttt{fallacy\_name: fallacy\_description}'' as described in Section~\ref{sec:fallacies-overview} \\
        \midrule
        \texttt{\$MACROCAT} & \faCheckSquare & \faCheckSquare & & \faCheckSquare & \makecell[cl]{\emph{Fallacies are divided into three categories:}\\\emph{- Insufficient Proof: Evading the Burden of Proof; Vagueness.}\\\emph{- Simplification: Hasty Generalization; Vagueness; False Dilemma; Slippery Slope; Causal}\\~~~~\emph{Oversimplification; Circular Reasoning; Thought-Terminating Cliché; Cherry Picking.}\\\emph{- Distraction: Red Herring; Cherry Picking; Appeal to Emotion; Thought-Terminating Cliché;}\\~~~~\emph{Slogan; Flag Waving; Loaded Language; Appeal to Authority; False Analogy; Strawman;}\\~~~~\emph{Ad Hominem; Name Calling or Labeling; Doubt.}} \\
        \midrule
        \texttt{\$INSTR\_1} & \faCheckSquare & \faCheckSquare & \faCheckSquare & & \makecell[cl]{\emph{You can only and exclusively use the labels I have listed for you. Remember to not modify the}\\~~~~\emph{names of the fallacies!}} \\
        \texttt{\$INSTR\_1} & \faCheckSquare & & & \faCheckSquare & \makecell[cl]{\emph{You must detect the portions of text that contain fallacies and classify them as Insufficient}\\~~~~\emph{Proof, Simplification and Distraction. Remember to only use these three labels!}} \\
        \texttt{\$INSTR\_1} & & \faCheckSquare &  & \faCheckSquare & \makecell[cl]{\emph{You must annotate the text with the categories Insufficient Proof, Simplification and Distraction.}\\~~~~\emph{Remember to only use these three labels!}} \\
        \midrule
        \texttt{\$INSTR\_2} & \faCheckSquare &  & \faCheckSquare & \faCheckSquare & \makecell[cl]{\emph{Fallacies can cover one or more tokens and they can overlap (more fallacies can be found on}\\~~~~\emph{the same segment of text). When detecting the span of text, try to respect linguistic units, for}\\~~~~\emph{example keep together consistent phrases or clauses, and when a logical passage includes}\\~~~~\emph{more sentences, annotate all the important information.}} \\
        \midrule
        \texttt{\$INSTR\_3} & & \faCheckSquare &  & \faCheckSquare & \makecell[cl]{\emph{A text can include up to three categories.}} \\
        \midrule
        \texttt{\$INSTR\_4} & \faCheckSquare &  & \faCheckSquare & \faCheckSquare & \makecell[cl]{\emph{, split into individual tokens with associated identification numbers. Maintain this tokenization}\\~~~~\emph{and do not alter the text}} \\
        \midrule
        \texttt{\$INPUT} & \faCheckSquare &  & \faCheckSquare & \faCheckSquare & The input text with one token per line following the format ``\texttt{token\_id [tab] token\_text}''\\
        \texttt{\$INPUT} &  & \faCheckSquare & \faCheckSquare & \faCheckSquare & The input text as it appears in its original form\\
        \midrule
        \texttt{\$INSTR\_5} & \faCheckSquare &  & \faCheckSquare & \faCheckSquare & \makecell[cl]{\emph{Your task is to detect the spans containing fallacies by indicating the first and last token}\\~~~~\emph{numbers, and then classify them into the \{twenty labels | three categories\} provided. It’s}\\~~~~\emph{really important that you DO NOT add any introduction, greetings, explanations,}\\~~~~\emph{descriptions and additional sentences. The output you will produce must follow the}\\~~~~\emph{format \texttt{[first number-last number = \{Fallacy Label | Category\}]}, for example:}\\~~~~\emph{\texttt{[1-6 = \{Evading the Burden of Proof | Insufficient Proof\}]}, for each identified}\\~~~~\emph{\{fallacy | fallacy category\}. If you do not find any fallacy, return \texttt{[None]}.}} \\
        \texttt{\$INSTR\_5} &  & \faCheckSquare & \faCheckSquare & \faCheckSquare & \makecell[cl]{\emph{It’s really important that you DO NOT add any introduction, greetings, explanations,}\\~~~~\emph{descriptions and additional sentences. The output you will produce must follow the format}\\~~~~\emph{\texttt{[\{Fallacy Label | Category\}]}, for example: \texttt{[\{Evading The Burden of Proof |}}\\~~~~\emph{\texttt{Insufficient Proof\}]}, for each identified \{fallacy | fallacy category\}. If you do not find}\\~~~~\emph{any fallacy, return \texttt{[None]}. }} \\
	\bottomrule
	\end{tabular}
        }%
	\caption{\label{tab:prompt-vars} Variables used in the prompt template above for \textsc{span} or \textsc{post} level tasks, using fine-grained (\textsc{f}) or coarse-grained (\textsc{c}) labels. In \texttt{\$INSTR\_5} we use the notation ``\{\textsc{f}|\textsc{c}\}'' to indicate alternatives for \textsc{f} and \textsc{c}, respectively.}
\end{table*}

\begin{table*}[!ht]
  \centering
  \resizebox{1.0\linewidth}{!}{%
  \begin{tabular}{clrra|rra|rra}
    \toprule
    & & \multicolumn{3}{c|}{\textbf{$\mathcal{A}_1 + \mathcal{A}_2$}} & \multicolumn{3}{c|}{\textbf{$\mathcal{A}_1$}} & \multicolumn{3}{c}{\textbf{$\mathcal{A}_2$}} \\
    & \textbf{Model} & \textbf{P} & \textbf{R} & \multicolumn{1}{r|}{\textbf{F$_1$}} & \textbf{P} & \textbf{R} & \multicolumn{1}{r|}{\textbf{F$_1$}} & \textbf{P} & \textbf{R} & \multicolumn{1}{r}{\textbf{F$_1$}} \\
    \midrule
    \multirow{4}{*}{\rotatebox[origin=c]{90}{\textsc{\textbf{POST-C}}}} &
    \textsc{MVML-alb} & 80.0$_{\pm1.5}$ & 74.0$_{\pm2.3}$ & \textbf{76.8}$_{\pm1.6}$ & 78.8$_{\pm1.9}$ & 70.6$_{\pm3.5}$ & \textbf{74.5}$_{\pm2.7}$ & 81.3$_{\pm1.0}$ & 77.3$_{\pm1.1}$ & \textbf{79.2}$_{\pm0.4}$ \\
    & \textsc{MVML-umb} & 84.5$_{\pm1.3}$ & 70.1$_{\pm4.2}$ & 76.6$_{\pm2.8}$ & 83.4$_{\pm1.2}$ & 67.2$_{\pm4.4}$ & 74.4$_{\pm3.1}$ & 85.6$_{\pm1.3}$ & 73.0$_{\pm3.9}$ & 78.8$_{\pm2.6}$ \\
    & \textsc{ZSWD-llama} & 57.9$_{\pm1.9}$ & 70.0$_{\pm1.9}$ & 63.3$_{\pm1.5}$ & 54.8$_{\pm2.0}$ & 69.3$_{\pm1.7}$ & 61.2$_{\pm1.7}$ & 60.9$_{\pm1.8}$ & 70.6$_{\pm2.1}$ & 65.4$_{\pm1.3}$ \\
    & \textsc{ZSWD-mixtr} & 64.7$_{\pm1.6}$ & 45.2$_{\pm1.0}$ & 53.2$_{\pm1.2}$ & 63.2$_{\pm1.7}$ & 46.2$_{\pm1.6}$ & 53.4$_{\pm1.6}$ & 66.1$_{\pm1.6}$ & 44.2$_{\pm0.5}$ & 53.0$_{\pm0.8}$ \\
    \midrule
    \multirow{4}{*}{\rotatebox[origin=c]{90}{\textsc{\textbf{POST-F}}}} &
    \textsc{MVML-alb} & 63.0$_{\pm2.0}$ & 34.3$_{\pm1.9}$ & \textbf{44.3}$_{\pm1.9}$ & 60.4$_{\pm1.7}$ & 31.2$_{\pm1.4}$ & \textbf{41.1}$_{\pm1.5}$ & 65.5$_{\pm2.2}$ & 37.4$_{\pm2.4}$ & \textbf{47.6}$_{\pm2.4}$ \\
    & \textsc{MVML-umb} & 39.0$_{\pm3.7}$ & 14.6$_{\pm1.6}$ & 21.3$_{\pm2.2}$ & 31.5$_{\pm4.6}$ & 11.5$_{\pm1.8}$ & 16.9$_{\pm2.6}$ & 46.5$_{\pm2.7}$ & 17.7$_{\pm1.5}$ & 25.7$_{\pm1.9}$ \\
    & \textsc{ZSWD-llama} & 20.9$_{\pm1.5}$ & 24.3$_{\pm2.3}$ & 22.5$_{\pm1.8}$ & 20.7$_{\pm1.4}$ & 24.7$_{\pm2.1}$ & 22.5$_{\pm1.6}$ & 21.1$_{\pm1.6}$ & 23.9$_{\pm2.5}$ & 22.5$_{\pm2.0}$ \\
    & \textsc{ZSWD-mixtr} & 26.0$_{\pm1.8}$ & 18.1$_{\pm1.4}$ & 21.4$_{\pm1.5}$ & 25.7$_{\pm1.9}$ & 18.4$_{\pm1.4}$ & 21.4$_{\pm1.6}$ & 26.3$_{\pm1.7}$ & 17.8$_{\pm1.3}$ & 21.3$_{\pm1.5}$ \\
    \midrule
    \multirow{4}{*}{\rotatebox[origin=c]{90}{\textsc{\textbf{SPAN-C}}}} &
    \textsc{MVMD-alb} & 55.2$_{\pm1.7}$ & 51.7$_{\pm2.1}$ & 53.3$_{\pm1.4}$ & 55.8$_{\pm1.2}$ & 50.3$_{\pm2.4}$ & 52.9$_{\pm1.8}$ & 54.6$_{\pm2.1}$ & 53.1$_{\pm1.8}$ & 53.8$_{\pm1.0}$ \\
    & \textsc{MVMD-umb} & 59.8$_{\pm1.5}$ & 50.4$_{\pm2.4}$ & \textbf{54.7}$_{\pm1.5}$ & 61.8$_{\pm1.5}$ & 49.9$_{\pm2.6}$ & \textbf{55.2}$_{\pm1.7}$ & 57.9$_{\pm1.4}$ & 50.8$_{\pm2.1}$ & \textbf{54.1}$_{\pm1.3}$ \\
    & \textsc{ZSWD-llama} & 25.3$_{\pm4.2}$ & 7.0$_{\pm0.8}$ & 10.9$_{\pm0.9}$ & 25.2$_{\pm4.9}$ & 6.9$_{\pm0.7}$ & 10.8$_{\pm1.0}$ & 25.4$_{\pm3.6}$ & 7.0$_{\pm0.9}$ & 10.9$_{\pm0.9}$ \\
    & \textsc{ZSWD-mixtr} & 31.6$_{\pm1.2}$ & 20.9$_{\pm1.4}$ & 25.1$_{\pm1.2}$ & 31.8$_{\pm1.3}$ & 20.9$_{\pm1.0}$ & 25.2$_{\pm0.9}$ & 31.3$_{\pm1.1}$ & 20.8$_{\pm1.7}$ & 25.0$_{\pm1.5}$ \\
    \midrule
    \multirow{12}{*}{\rotatebox[origin=c]{90}{\textsc{\textbf{SPAN-F}}}} &
    \textbf{\emph{Strict mode}} &  &  &  &  &  & \\
    & ~~~\textsc{MVMD-alb} & 47.6$_{\pm1.9}$ & 25.6$_{\pm1.6}$ & \textbf{33.3}$_{\pm1.4}$ & 47.1$_{\pm2.1}$ & 24.4$_{\pm1.6}$ & \textbf{32.1}$_{\pm1.3}$ & 48.0$_{\pm1.7}$ & 26.8$_{\pm1.5}$ & \textbf{34.4}$_{\pm1.4}$ \\
    & ~~~\textsc{MVMD-umb} & 57.5$_{\pm5.9}$ & 3.9$_{\pm0.7}$ & 7.3$_{\pm1.3}$ & 51.7$_{\pm4.8}$ & 5.0$_{\pm0.9}$ & 9.2$_{\pm1.5}$ & 63.3$_{\pm6.9}$ & 2.9$_{\pm0.6}$ & 5.5$_{\pm1.0}$ \\
    & ~~~\textsc{ZSWD-llama} & 4.5$_{\pm0.5}$ & 2.7$_{\pm0.4}$ & 3.4$_{\pm0.3}$ & 4.5$_{\pm0.4}$ & 2.6$_{\pm0.4}$ & 3.3$_{\pm0.4}$ & 4.5$_{\pm0.6}$ & 2.8$_{\pm0.3}$ & 3.4$_{\pm0.3}$ \\
    & ~~~\textsc{ZSWD-mixtr} & 5.8$_{\pm1.1}$ & 3.2$_{\pm0.5}$ & 4.2$_{\pm0.7}$ & 5.9$_{\pm1.1}$ & 3.3$_{\pm0.6}$ & 4.2$_{\pm0.7}$ & 5.7$_{\pm1.0}$ & 3.2$_{\pm0.4}$ & 4.1$_{\pm0.6}$ \\
    & \textbf{\emph{Soft mode}} &  &  &  &  &  & \\
    & ~~~\textsc{MVMD-alb} & 52.2$_{\pm2.0}$ & 28.7$_{\pm1.7}$ & \textbf{37.0}$_{\pm1.5}$ & 52.0$_{\pm2.1}$ & 27.7$_{\pm1.8}$ & \textbf{36.1}$_{\pm1.5}$ & 52.3$_{\pm1.8}$ & 29.7$_{\pm1.6}$ & \textbf{37.9}$_{\pm1.5}$ \\
    & ~~~\textsc{MVMD-umb} & 66.3$_{\pm5.5}$ & 4.8$_{\pm0.7}$ & 8.9$_{\pm1.3}$ & 65.3$_{\pm3.6}$ & 6.4$_{\pm0.9}$ & 11.7$_{\pm1.6}$ & 67.2$_{\pm7.4}$ & 3.1$_{\pm0.5}$ & 6.0$_{\pm0.9}$ \\
    & ~~~\textsc{ZSWD-llama} & 6.4$_{\pm0.6}$ & 4.2$_{\pm0.5}$ & 5.0$_{\pm0.4}$ & 6.4$_{\pm0.8}$ & 4.0$_{\pm0.6}$ & 4.9$_{\pm0.7}$ & 6.4$_{\pm0.4}$ & 4.4$_{\pm0.4}$ & 5.2$_{\pm0.2}$ \\
    & ~~~\textsc{ZSWD-mixtr} & 8.2$_{\pm1.5}$ & 5.4$_{\pm1.0}$ & 6.5$_{\pm1.1}$ & 8.3$_{\pm1.6}$ & 5.5$_{\pm1.1}$ & 6.6$_{\pm1.2}$ & 8.1$_{\pm1.5}$ & 5.3$_{\pm0.8}$ & 6.4$_{\pm1.1}$ \\
    \bottomrule
    
  \end{tabular}
  }%
  \caption{\label{tab:results-individual} Test set results for \textsc{post} and \textsc{span} tasks at the \emph{coarse-grained} (\textsc{c}) and \emph{fine-grained} (\textsc{f}) level. We report average precision (P), recall (R), and F$_1$ scores (w/ std dev) across $k=5$ splits, both on individual test sets (i.e., $\mathcal{A}_1$ and $\mathcal{A}_2$) and averaged over all $|\mathcal{A}|$ test set versions (i.e., $\mathcal{A}_1 + \mathcal{A}_2$). For the \textsc{span-f} task, we also present scores using both \emph{strict} and \emph{soft} modes. Best results for each task are in bold.}
\end{table*}

\end{document}